\documentclass[preprint,12pt]{elsarticle}

\usepackage{amsmath, graphicx}
\usepackage{amsfonts, amsthm, amssymb, mathrsfs, mathtools}
\usepackage{bm}
\usepackage{algorithm, algorithmic}
\usepackage{setspace}
\usepackage{subfig}
\usepackage{array}
\usepackage{booktabs}
\usepackage{subfiles}
\usepackage{multirow}
\usepackage{cases}
\usepackage{makecell}
\usepackage{bbding}
\usepackage{footmisc}
\usepackage{rotating}
\usepackage{caption}
\usepackage{ulem}
\usepackage{diagbox}

\usepackage[english]{babel}
\usepackage{url}
\usepackage[colorlinks,
linkcolor=magenta,
anchorcolor=magenta,
citecolor=magenta,
]{hyperref}
\newcommand{\N}{\mathcal{N}}
\newcommand{\X}{\mathcal{X}}
\newcommand{\Y}{\mathcal{Y}}
\newcommand{\I}{\mathcal{I}}
\newcommand{\A}{\mathcal{A}}
\newcommand{\E}{\mathbb{E}}
\newcommand{\R}{\mathbb{R}}
\newcommand{\brm}[1]{\bm{\mathrm{#1}}}
\newcommand{\Down}{\bm{\mathrm{D}}}
\newcommand{\Blur}{\bm{\mathrm{B}}}
\newcommand{\sr}{\bm{\mathrm{r}}}
\newcolumntype{M}[1]{>{\centering\arraybackslash}m{#1}}
\newcommand{\tb}[1]{\textbf{#1}}
\newcommand{\ul}[1]{\underline{#1}}
\newcommand{\w}{\bm{\mathrm{w}}}

\journal{Information Fusion}

\begin{document}

\begin{frontmatter}



\title{Unsupervised Hyperspectral Pansharpening via Low-rank Diffusion Model}
\author[1]{Xiangyu Rui\fnref{2}}
\ead{xyrui.aca@gmail.com}

\author[1]{Xiangyong Cao\corref{cor1}\fnref{3}}
\ead{caoxiangyong@xjtu.edu.cn}

\author[1]{Li Pang\fnref{3}}
\ead{2195112306@stu.xjtu.edu.cn}

\author[1]{Zeyu Zhu\fnref{4}}
\ead{zeyuzhu2077@gmail.com}

\author[1]{Zongsheng Yue\fnref{2}}
\ead{zsyzam@gmail.com}

\author[1]{Deyu Meng\fnref{2}\corref{cor1}}
\ead{dymeng@mail.xjtu.edu.cn}

\fntext[2]{Xiangyu Rui, Zongsheng Yue and Deyu Meng are with the School of Mathematics and Statistics and Ministry of Education Key Lab of Intelligent Network Security.}
\fntext[3]{Xiangyong Cao and Li Pang are with the School of Computer Science and Technology and Ministry of Education Key Lab For Intelligent Networks and Network Security.}
\fntext[4]{Zeyu Zhu is with the College of Artificial Intelligence.}
\cortext[cor1]{Corresponding author}

\affiliation[1]{organization={Xi'an Jiaotong University},
            addressline={Xianning West Road},
            city={Xi'an},
            postcode={710049},
            state={Shaanxi},
            country={China}}

            % country={}}

% \author[label1]{Xiangyu Rui, Xiangyong Cao, Li Pang, Zeyu Zhu, Zongsheng Yue, and Deyu Meng\corref{corauthor}.}

% \affiliation[label1]{organization={Xi'an Jiaotong University},%Department and Organization
%             city={Xi'an},
%             postcode={710049}, 
%             state={Shaanxi},
%             country={China}}
        
% \cortext[corauthor]{corresponding author}

\begin{abstract}
%Hyperspectral pansharpening is a process of merging a high-resolution panchromatic (PAN) image and a low-resolution hyperspectral (LRHS) image to create a single high-resolution hyperspectral (HRHS) image. Existing deep learning-based pansharpening methods have poor generalization ability and the traditional model-based pansharpening methods need careful manual exploration for the image structure prior. To alleviate these issues, this paper proposes an unsupervised pansharpening method by combining the diffusion model with the low-rank matrix factorization technique. Specifically, we assume that the HRHS image is decomposed into the product of two low-rank tensors, i.e., the base tensor and the coefficient matrix. The base tensor lies on the image field and has low spectral dimension, we can thus conveniently utilize a pre-trained remote sensing diffusion model to capture its image structures. Additionally, we derive a simple yet quite effective way to pre-estimate the coefficient matrix from the observed LRHS image, which preserves the spectral information of the HRHS. Extensive experimental results on some benchmark datasets demonstrate that our proposed method performs better than traditional model-based approaches and has better generalization ability than deep learning-based techniques. The code is released in \href{https://github.com/xyrui/PLRDiff}{https://github.com/xyrui/PLRDiff}. 

Hyperspectral pansharpening is a process of merging a high-resolution panchromatic (PAN) image and a low-resolution hyperspectral (LRHS) image to create a single high-resolution hyperspectral (HRHS) image. Existing Bayesian-based HS pansharpening methods require designing handcraft image prior to characterize the image features, and deep learning-based HS pansharpening methods usually require a large number of paired training data and suffer from poor generalization ability. To address these issues, in this work, we propose a low-rank diffusion model for hyperspectral pansharpening by simultaneously leveraging the power of the pre-trained deep diffusion model and better generalization ability of Bayesian methods. Specifically, we assume that the HRHS image can be recovered from the product of two low-rank tensors, i.e., the base tensor and the coefficient matrix. The base tensor lies on the image field and has a low spectral dimension. Thus, we can conveniently utilize a pre-trained remote sensing diffusion model to capture its image structures. Additionally, we derive a simple yet quite effective way to pre-estimate the coefficient matrix from the observed LRHS image, which preserves the spectral information of the HRHS. Experimental results demonstrate that the proposed method performs better than some popular traditional approaches and gains better generalization ability than some DL-based methods. The code is released in \href{https://github.com/xyrui/PLRDiff}{https://github.com/xyrui/PLRDiff}. 
\end{abstract}


%Research highlights
% \begin{highlights}
% \item An unsupervised method named PLRDiff is proposed to handle hyperspectral pansharpening problem.
% \item The PLRDiff utilizes a pre-trained unconditional diffusion model to better capture the hyperspectral image prior.
% \item An easy-to-implement strategy is proposed to project the HRHS image into a low-rank subspace for efficient sampling.
% \item The PLRDiff can finely generalize to different datasets and degradation conditions.
% \end{highlights}


\end{frontmatter}


\section{Introduction}
Optical remote sensing image, e.g., hyperspectral (HS) image, can record a wide range of spectrum information and has been widely used in many practical scenes, such as agriculture~\cite{HSI-app-agricultural1}, geology~\cite{HSI-app-geology} and food industry~\cite{HSI-app-food}. To sufficiently explore the data information, HS image with high spatial resolution and high spectral resolution is demanded. However, remote sensors cannot simultaneously obtain high-spatial resolution and high-spectral resolution HS images due to the technique limitation. As a result, low-spatial resolution hyperspectral images with high spectral resolution (LRHS) and high-spatial-resolution panchromatic (PAN) images with single spectral dimension are commonly collected instead. Therefore, it is a crucial task to generate a high-spatial and high-spectral resolution hyperspectral (HRHS) image by fusing the corresponding LRHS image and PAN image. This fusion task is called hyperspectral pansharpening in the remote sensing field and has been studied for decades. Another simlilar task is called multispectral (MS) pansharpening, where the observed low spatial resolution image is an MS image that has a much lower spectral dimension (usually less than 10) than the HS image. As depicted in \cite{Hyperspectral-pansharpening-A-review}, the broader spectrum range of the HS image than that of MS image brings more difficulty for HS pansharpening. 

Previous works on solving the HS pansharpening problem can be mainly categorized into four classes~\cite{cao2022proximal}, i.e. the component substitution (CS) based methods, the multiresolution analysis (MRA) based methods, the Bayesian methods and the deep learning (DL) based methods. CS-based methods and MRA-based methods are originally designed for MS pansharpening but can also be applied to HS pansharpening \cite{Hyperspectral-pansharpening-A-review}. Specifically, the CS-based methods transform the LRHS image into another space that is assumped to separate the spatial and spectral information. The transformed image will then be enhanced by the information from the PAN image and the spatial structures from the transformed image. The MRA-based methods inject the spatial information that is extracted from the PAN image into the resized LRHS image. The Bayesian methods formulate the pansharpening task from a statistical perspective. Taking the LRHS and PAN images as conditions, Bayesian methods involve the inference of the posterior distribution of the HRHS image. Along this line, the variational optimization (VO) based methods formulate an optimization problem, which can be equivalently seen as maximizing a posterior probability. The performance of Bayesian methods largely depends on how the image prior is characterized. In recent years, DL-based methods have emerged as a promising approach for pansharpening because of their remarkable performance. These methods utilize the powerful non-linear fitting ability of deep neural networks to directly learn the mapping from LRHS and PAN images to HRHS image by feeding a large number of training pairs. However, the DL-based methods usually lack good generalization ability due to the distribution shift of training and test data. Therefore, DL-based methods tend to be fragile when faced with changeable conditions.

%The Bayesian methods regard pansharpening as an inverse problem and construct a corresponding objective function that describes the relationship among the desired HRHS image, the LRHS image, and the PAN image. HRHS image is then derived by optimizing the objective function through a properly selected algorithm. Their performance largely rely on the image prior characterization.

In this work, we seek to balance the good generalization ability of Bayesian methods and the advanced data fitting ability of DL-based methods and thus propose an unsupervised low-rank diffusion model (called PLRDiff) for the HS pansharpening task. Specifically, the deep networks are used to find the complex prior information of HS images, which is independent of the specific degradation process. In order to fully exploit the HS image prior, unlike the Bayesian methods, we resort to the powerful deep generative model that could directly learn the data distribution. Popular generative models include variational autoencoder (VAE)~\cite{kingma2013auto}, generative adversarial network (GAN)~\cite{goodfellow2020generative}, flow model~\cite{dinh2014nice} and diffusion model~\cite{DDPM}. Among these generative models, the recently proposed diffusion model has the advantage of stable training and excellent image generation ability~\cite{rombach2022high}. The diffusion model could sample from the data distribution by gradually sampling along a Markov chain that starts from a known simple distribution and ends to the data distribution. Furthermore, recent work \cite{SDE} further reveals the relationship between diffusion models and stochastic differential equations, making it flexible to combine external observation information with the diffusion model \cite{Improving-diffusion-models-for-inverse-problems-using-manifold-constraints, Diffusion-Model-Based-Posterior-Sampling-for-Noisy-Linear-Inverse-Problems, Solving-inverse-problems-in-medical-imaging-with-score-based-generative-models}. Thus, we could conveniently apply the diffusion model to describe the HS image prior and restore the HRHS image with the guidance of the LRHS image and the PAN image.

As mentioned above, an HS image could contain hundreds of bands and its spectral dimension varies across datasets. This brings about two problems if we use the diffusion model to learn the entire HS image distribution. Firstly, when the spectral dimension is very large, it is very difficult to learn the image distribution and the time cost is also very high. Secondly, when the spectrum covers a wide range, it is difficult to collect enough HS images for the diffusion model to efficiently learn the image distribution. Thus, instead of sampling from the original HS image distribution, we consider projecting the HS image into a lower-dimensional subspace and performing sampling in this subspace. When the sampling is done, we project the sample back into the original HS image field. 

Since the HS image has low-rank property along its spectral dimension~\cite{NGmeet, Fasthy, cao2016robust}, it is natural to represent the HS image by the product of two tensors. One of them has the same spatial dimensions as the HS image but has a much lower spectral dimension, which is called the base tensor. In this work, we aim to inference the base tensor by the pre-trained diffusion model. The other one is a coefficient matrix that describes how the HS image is spanned by the base tensor. It can be easily seen that the coefficient matrix mainly contains the spectral information of an HS image. Obviously, such factorization form is not unique. In order to efficiently use a pre-trained diffusion model to capture the distribution of the base tensor, in this work, we factorize the HS image in a self-representation way. Specifically, we formulate the base tensor by the linearly independent bands from the HS image itself. Under such factorization form, we find that the HRHS image and the LRHS image share the same coefficient matrix and thus it could be conveniently estimated from the observed LRHS image at first. 

%The coefficient matrix is used to project the HRHS image into the low-dimension subspace. Then we further utilize the diffusion model to efficiently sample the base tensor from this subspace. Finally, we use the learnt base tensor and the pre-estimated coefficient matrix to restore the HRHS image.  

%Obviously, such factorization is not unique. Typical ways include principle component analysis (PCA), sparse representation and non-negative matrix factorization (NMF). However, the base tensor calculated by these ways are not similar to those of real images. Thus, we turn to a

%and the feature of base tensor generated by common decomposition strategies, such as principle component analysis (PCA) and sparse representation, is not similar to those of real images. In this way, it is not convenient to explicitly utilize the well-studied knowledge of image distribution.  
In summary, our contributions are as follows:
\begin{itemize}
	\item We propose an unsupervised pansharpening method (PLRDiff) that utilizes the low-rank property of HS image and the unconditional diffusion model. The diffusion model has the strong power to directly learn the image distribution, which is irrelevant to any degradations. The image degradation information can be additionally encoded in the sampling process, which makes the proposed method flexible to handle different degradations. 
	\item We propose an easy-to-implement projection strategy for an efficient generation. Specifically, we consider projecting the HRHS image into a subspace base on the low-rank property of HS image. The subspace has a lower spectral dimension so we could conveniently and effectively perform the sampling in this subspace. 
	\item Our PLRDiff method is unsupervised and has a natural ability to generalize. Thus it is flexible to be applied to different datasets and degradation conditions. Extensive experiments show that our PLRDiff method achieves better performance than traditional model-based approaches and has better generalization ability than DL-based approaches.
\end{itemize}

This work is organized as follows. In Sec. \ref{sec-related-work}, we review the previous HS pansharpening methods. In Sec. \ref{notation}, we define some notations. In Sec. \ref{sec-main-method}, we present the proposed unsupervised low-rank diffusion model for HS pansharpening in detail. In Sec. \ref{sec-experiments}, we conduct several experiments on different datasets to verify the effectiveness of the proposed method. In Sec. \ref{sec-discussion}, we make some discussions for our method. 

\section{Related Works}\label{sec-related-work}
In this section, we review four categories of pansharpening methods, i.e. CS-based methods, MRA-based methods, Bayesian methods and DL-based methods.

The CS-based methods are implemented by taking a transformation on the LRHS image and then substituting the spatial components from the transformed LRHS image with information extracted from the PAN image to enhance the image quality. The transformation is assumed to be able to separate the spatial and spectral structures of the LRHS image. Typical CS-based methods include Intensity-hue-saturation (IHS) \cite{IHS1, IHS2}, Brovey transformation \cite{Brovey}, Band-Dependent Spatial-Detail (BDSD) \cite{BDSD}, Gram-Schmidt (GS) \cite{GS}, Partial Replacement Adaptive Component Substitution (PRACS) \cite{PRACS}, Indusion \cite{Indusion}, etc. The main difference between these methods is the way of comparing the components of the transformed LRHS image with the histogram-matched PAN image. The CS-based methods are easy to implement and enjoy low time expense. However, they may suffer from spectral information distortion when the information from the PAN image and the transformed image is not finely aligned \cite{Review-VG}.

The MRA-based methods consider injecting the spatial information extracted from the PAN image into the rescaled LRHS image. Specifically, the details are derived by comparing the PAN image with its filtered version. For example, the high-pass filtering (HPF) method \cite{HPF} 
takes box filtering on the histogram-matched PAN image and extracts details by the difference between the PAN image and the filtered image. The smoothing filter-based intensity modulation (SFIM) method \cite{SFIM} also uses box filtering, however, the rescaled LRHS image is multiplied by the detail information rather than added to. Along this research line, more transforms on the LRHS and PAN images and filtering strategies are explored, such as wavelet and contourlet decomposition. Typical methods include Indusion \cite{Indusion}, MTF-GLP \cite{MTF-GLP}, etc. Compared with CS-based method, MRA-based methods could restore more spectral information but may reduce the spatial quality \cite{A-critical-comparison-among-pansharpening-algorithms}.  

The Bayesian methods consider inferencing the posterior distribution of the HRHS image. They usually require to model the HS image prior and the degradation process. For example, \cite{Bayesian-fusion-of-multi-band-images} assigns a Gaussian prior to the projected HRHS image and performs hybrid Gibbs sampling to get the restored HRHS image. Within this class, the variational optimization methods solve an optimization problem that is equivalent to maximum a posterior (MAP) estimation. Then appropriate optimization algorithms are applied to derive the optimal HRHS image. P+XS method \cite{P+XS} assumes that the LRHS image is blurred and downsampled by the HRHS image, and the PAN image can be linearly represented by the HRHS image along the spectral dimension. HySure \cite{HySure} factorize the HRHS image into the product of a base matrix and a coefficient matrix by singular value decomposition (SVD). The gradients of the base matrix are regularized to constrain the restored HRHS image. Many other VO based methods are more popular for MS pansharpening \cite{A-variational-approach-for-pan-sharpening, FU2019, A-new-pansharpening-method-based-on-spatial-and-spectral-sparsity-priors,The-fusion-of-panchromatic-and-multispectral-remote-sensing-images-via-tensor-based-sparse-modeling-and-hyper-Laplacian-prior, vo+, LRTCF}. Additionally, a matrix factorization method, i.e., CNMF \cite{CNMF}, performs coupled nonnegative matrix factorization on the HRHS image and sequentially updates the components. The Bayesian methods could carefully model complex HS image prior and the degradation. They usually have good generalization ability and interpretability to the restoration process. But the HS prior is manually designed and may not fully capture the image structure.

The DL-based methods for pansharpening have been extensively studied recently \cite{Review-DLJ}. This line of work benefits from sufficient training pairs and advanced network architectures. Many popular works are initially designed for MS pansharpening \cite{PNN,MSDCNN,PANnet,FusionNet,SRPPNN,PANcsc, cao2022proximal}. For HS pansharpening, \cite{HyperPNN} introduce a spectrally predictive structure to enhance the spectral structures of the HRHS image. \cite{Laplacian-pyramid-dense-network-for-hyperspectral-pansharpening} proposes a deep network that is a cascade of the Laplacian pyramid dense networks.
\cite{Multistage-dual-attention-guided-fusion-network-for-hyperspectral-pansharpening} propose a network that contains a three-stream structure, a dual-attention guided fusion block and a multiscale residual dense block to fully exploit the image structures. \cite{ Generative-dual-adversarial-network-with-spectral-fidelity-and-spatial-enhancement-for-hyperspectral-pansharpening} use a generative adversarial network to solve the pansharpening problem. Two discriminators are used to preserve the spatial features and spectral features, respectively. \cite{Hypertransformer-A-textural-and-spectral-feature-fusion-transformer-for-pansharpening} utilize the transformer structure to learn the cross-feature
space dependencies and long-range details of the PAN and
LRHS image. \cite{A-deep-shallow-fusion-network-with-multidetail-extractor-and-spectral-attention-for-hyperspectral-pansharpening} proposes a network that contains a multi-detail extractor, a deep-shallow fusion structure and a spectral attention module to better extract the spectral and spatial features. The above works are under supervised learning manner and usually need to retrain or finetune the networks for different datasets. Some works turn to unsupervised learning manner.
\cite{Spectral-fidelity-convolutional-neural-networks-for-hyperspectral-pansharpening} introduces two HSpeNet methods and learns the network by a spectral-fidelity loss. 
\cite{Hyperspectral-pansharpening-using-deep-prior-and-dual-attention-residual-network} propose a method using deep hyperspectral prior and dual-attention residual networks. \cite{DDRF} is the first work that utilizes the diffusion model. However, it is still in a supervised learning manner that requires a large amount of training pairs. In this paper, we first propose an unsupervised HS pansharpening method based on the pre-trained diffusion model as far as we know.

\section{Notations}\label{notation}
$\X\in\R^{H\times W\times S}$ represents a third order tensor. The $i$th slice of $\X$ is a matrix denoted by $\X(:,:,i)$. $\brm{X}_{(3)}\in\R^{S\times HW}$ means the mode-3 unfolding of $\X$ by flattening the first two dimensions. On the contrary, $\textbf{fold}(\brm{X}_{(3)})$ means reshaping $\brm{X}_{(3)}$ back to $\X$. ``$\X\times_3 E$" is the mode-3 tensor multiplication between $\X$ and a matrix $E$:
\begin{align}
	\X\times_3 E = \textbf{fold}(E\times \brm{X}_{(3)}). \nonumber
\end{align}
A matrix is also considered as a second-order tensor. ``$E^T$" represents the transpose matrix of $E$. ``$||\X||_F$" represents the Frobenius norm of $\X$. ``$[N]$" means $\{1,2,...,N\}$.

\section{Main Method}\label{sec-main-method}
\begin{figure}[t]
	\centering
	\includegraphics[width=13.5cm]{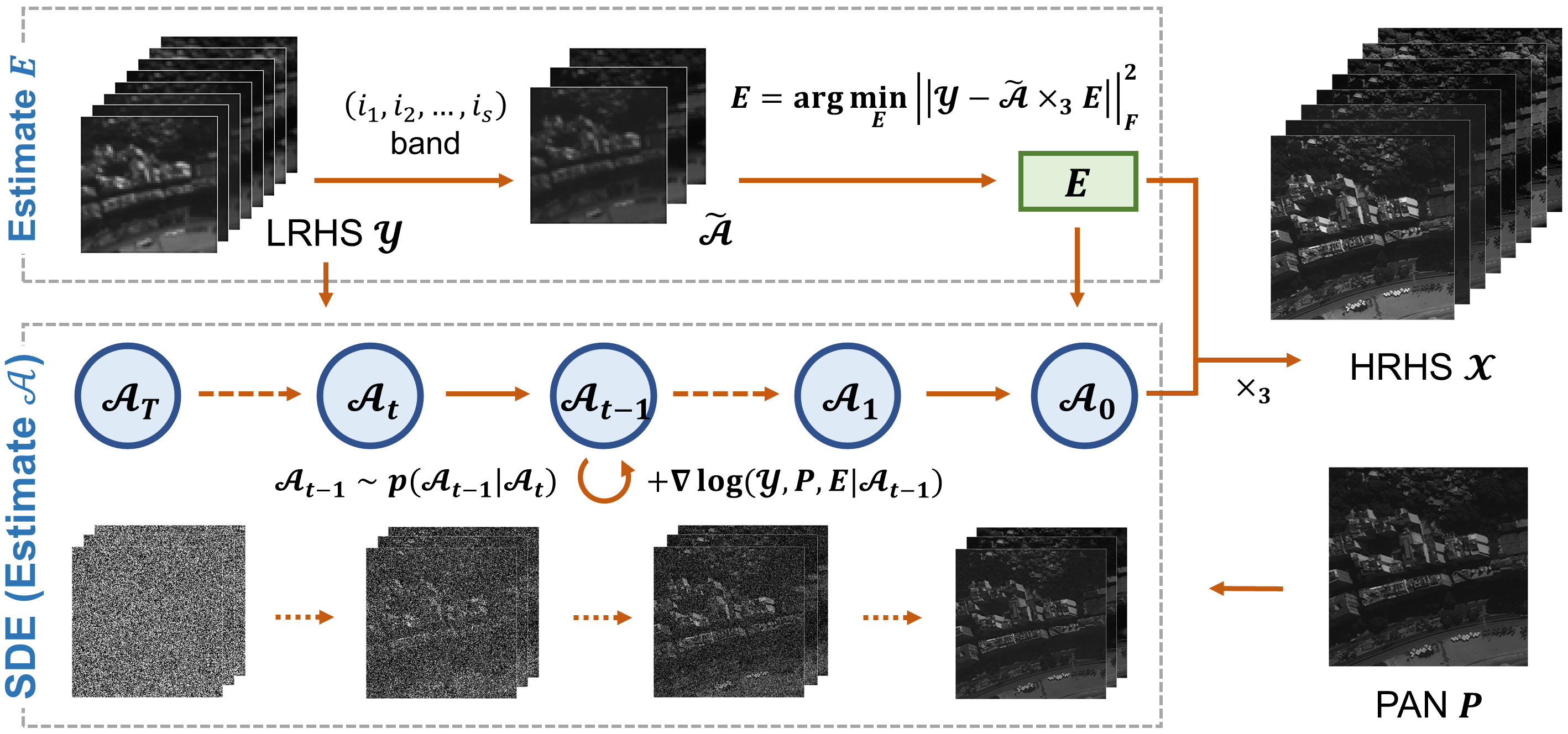}
	\caption{The flowchart of the proposed method (PLRDiff). First, we estimate the coefficient matrix $E$ from the LRHS image. Second, the LRHS image, PAN image and the coefficient matrix $E$ served as conditions are sent into a discretization of an SDE, where we reconstruct the base tensor $\A$ by gradually sampling through a series of parameterized distributions. Finally, the predicted $\A$ and $E$ are multiplied to form the desired HRHS image.}
	\label{fig-main}
\end{figure}

Pansharpening is a kind of image restoration task that aims to construct an HRHS image $\X\in\R^{H\times W\times S}$ from the observed LRHS image $\Y\in\R^{h\times w\times S}$ and the PAN image $P\in\R^{H\times W}$. The LRHS image $\Y$ has lower spatial resolution compared with the HRHS image $\X$. Usually, it is assumed that the LRHS image is blurred and downsampled by the HRHS image \cite{HySure}\cite{CNMF}, which can be formulated as follows  
\begin{align}\label{spectral-constraint}
	\Y  = \Down(\Blur(\X)).
\end{align}
The operator $\Down$ and $\Blur$ represent downsampling and blurring in the spatial dimension, respectively. The process of downsampling and blurring mainly damages the spatial structure of the HRHS image while maintaining its spectral information. The relationship between the HRHS image and the PAN image has several descriptions. A common one of them is that the PAN image can be linearly represented by the HRHS image along the spectral dimension \cite{P+XS}\cite{The-fusion-of-panchromatic-and-multispectral-remote-sensing-images-via-tensor-based-sparse-modeling-and-hyper-Laplacian-prior}\cite{An-interpretable-unsupervised-unrolling-network-for-hyperspectral-pansharpening}, which can be written as  
\begin{align}\label{spatial-constraint}
	P = \X\times_3\sr.
\end{align}
``$\sr\in\R^{1\times S}$" represents the spectral response vector. The PAN image largely preserves the spatial information of the HRHS image while it losses much spectral information.

Unfortunately, we are still not able to restore the exact HRHS image only from the degradation processes (\ref{spectral-constraint}) and (\ref{spatial-constraint}), because the problem is heavily ill-posed, i.e., the solutions are not unique. Following the idea of the Bayesian methods, an image prior from the HS image field is needed to better restore the HRHS image. In other words, the prior helps to find the solution that lies in the HS image field as well as following (\ref{spectral-constraint}) and (\ref{spatial-constraint}). Previous Bayesian methods design the image prior manually, e.g., Gaussian prior \cite{Bayesian-fusion-of-multi-band-images}, which may not fully investigate the intrinsic image structures. Recently, the diffusion model has gained great success in generating high-quality images~\cite{DPS}\cite{stablediffusion} from complex image distribution $p(\X)$. Specifically, the diffusion model introduces a series of latent variables that gradually change the unknown image distribution $p(\X)$ into a simple known distribution $p_T(\X_T)$. Then, a sample from $p(\X)$ could be derived by sequentially sampling from $p_T(\X_T)$ back to $p(\X)$. In this work, we consider utilizing the existing \textit{pre-trained unconditional} diffusion model~\cite{DDPMCD} for remote sensing images to characterize the image distribution $p(\X)$. Unlike \cite{DDRF}, the unconditional diffusion model is only related to $p(\X)$ and is irrelevant to specific degradation processes. Thus, the proposed method should be more flexible when faced with different situations and easy to generalize. 

An essential problem for applying the diffusion model to HS image is that the HS image usually covers a broad spectrum range, which means ``$S$" is usually very large and changes from image to image. The high data dimensionality would bring unavoidable difficulties in efficiently learning the data distribution. Besides, it also requires much more computational resources and time to sample from such distribution. Thus, we consider projecting the HS image into a lower-dimensional subspace. Fortunately, it is widely known that the HS image has low-rank property along its spectral dimension, which means that a small amount of rank-1 matrices could linearly represent the mode-3 unfolding of an HS image while keeping its major information. In other words, the HS image can be decomposed into two low-rank tensors. One of them is the base tensor where the HS image is spanned from and the other is the coefficient matrix that means how the HS image is spanned from the base tensor. Assuming that the pre-trained diffusion model learns the distribution of images that have a lower spectral dimension, we consider formulating the base tensor from the HS image itself. In the rest of this section, we will first introduce the proposed low-rank decomposition, and then review the diffusion model. Finally, we present the overall proposed method.

\subsection{Low-rank Factorization}\label{sec-low-rank-factorization}
%Linear subspace representation (LSR) \cite{Pattern-classification, Linear-Subspace-Learning-Based-Dimensionality-Reduction} is a powerful tool for dimension reduction. Some typical LSR methods, e.g., PCA \cite{PCA} and sparse representation \cite{Sparse-representation} are widely used. Let $U=[u_1^T,u_2^T,...,u_S^T]^T$ be a data matrix containing $S$ observed data $u_i$, the LSR assumes that each $u_i$ can be linearly represented by another series of vector $\{v_i\}_{i=1}^s$: 
%\begin{align}
%	u_i = e_{i1}v_1 + e_{i2}v_2 + ... + e_{is}v_s,~i\in[S]. 
%\end{align}
%Let $V =[v_1^T,v_2^T,...,v_s^T]^T$, the above formulation can also be written as 
%\begin{align}
%	U = EV,
%\end{align}
%where $e_{ij}$ is the $(i,j)$th element of matrix $E$.

As mentioned above, the HS image $\X$ has low-rank property along its spectral dimension, which means $\X$ can be represented by 
\begin{align}\label{subspace-represent}
	\X = \A\times_3 E \Leftrightarrow \brm{X}_{(3)} = E\brm{A}_{(3)},
\end{align}
where $\brm{A}_{(3)}\in\R^{s\times HW}$ and $E\in\R^{S\times s}$ are low rank tensors, i.e., $s\ll S$. Here, $\brm{A}$ is assumed to still lie in the image field. To achieve this goal, a direct idea is to select $s$ linearly independent rows from $\brm{X}_{(3)}$ itself and constitute $\brm{A}_{(3)}$. Specifically, $\brm{A}_{(3)}$ has a concise form of 
\begin{align}\label{fac-a}
\brm{A}_{(3)} = [x_{i_1}^T,x_{i_2}^T,...,x_{i_s}^T]^T,
\end{align} 
where $x_{i_j}$ means the $i_j$-th row of $\brm{X}_{(3)}$. (\ref{fac-a}) also means that the $j$-th slices of $\A$ (i.e., $\A(:,:,j)$) is the $i_j$-th slices of $\X$ (i.e., $\X(:,:,i_j))$. In this way, we derive a low-rank factorization of $\X$ with an image-like $\A$ and a non-trivial coefficient $E$. $\X$ can be seen as lying in the subspace spanned by $\A$. The coefficient matrix $E$ mainly preserves the spectral information of $\X$. Given $\A$, $E$ can be estimated by 
\begin{align}\label{estimate-E-ori}
	E = \arg\min_E~\left\|\X - \A\times_3 E \right\|_F^2.
\end{align}

%Compare with PCA, the base tensor $\A$ by the proposed method lies on the original image field and preserves more image features. One observation is that this image field has very similar spatial features to those of the RGB image field, as shown in Fig. \ref{fig-sr-svd}.

For the pansharpening task, only the LRHS image $\Y$ and the PAN image $P$ are observed. We do not have access to the HRHS image $\X$ directly. However, we find that $\Y$ and $\X$ share the same coefficient matrix:
\begin{align}
	\Y & = \Down\left( \Blur\left( \X \right) \right)  \nonumber\\
	& = \Down(\Blur(\A\times_3 E))  \nonumber\\
	& = (\Down(\Blur(\A)))\times_3 E  \nonumber\\
	& := \tilde{\A}\times_3 E.
\end{align}
The above relationship is based on the assumption that the downsampling and blurring operators are linear, which generally holds. Looking closer to $\tilde{\A} = \Down(\Blur(\A))$, this means that $\tilde{\A}$ can be derived by bluring and downsampling $\A$. Since $\A$ is composed of slices of $\X$, $\tilde{\A}$ is exactly composed of slice of $\tilde{Y}$ with the same index $(i_1,...,i_s)$ as in (\ref{fac-a}). That is 
\begin{align}
	\tilde{\A}(:,:,j) = \Y(:,:,i_j), \forall j\in[s].
\end{align}

Thus, to estimate $E$, we just need to firstly select the $(i_1,i_2,...i_s)$-th slices of $\Y$ to formulate $\tilde{\A}$. And then $E$ is calculated by
\begin{align}\label{estimate-E}
	E = \arg\min_E~ \left\| \Y - \tilde{\A}\times_3 E \right\|_F^2.
\end{align}
In this work, the index $(i_1,i_2,...i_s)$ is selected at equal intervals along the spectral dimension of $\Y$. That is,
\begin{align}
	i_j = \left\lceil \dfrac{S}{s+1} \right\rceil\times j,~j\in[s].
\end{align}

Once the coefficient matrix $E$ is estimated, we then focus on restoring the base tensor $\A$.

\subsection{Denoising Diffusion Probabilistic Model}\label{sec-denoising-diffusion-model}
\begin{figure}[t]
	\centering
	\includegraphics[width=10cm]{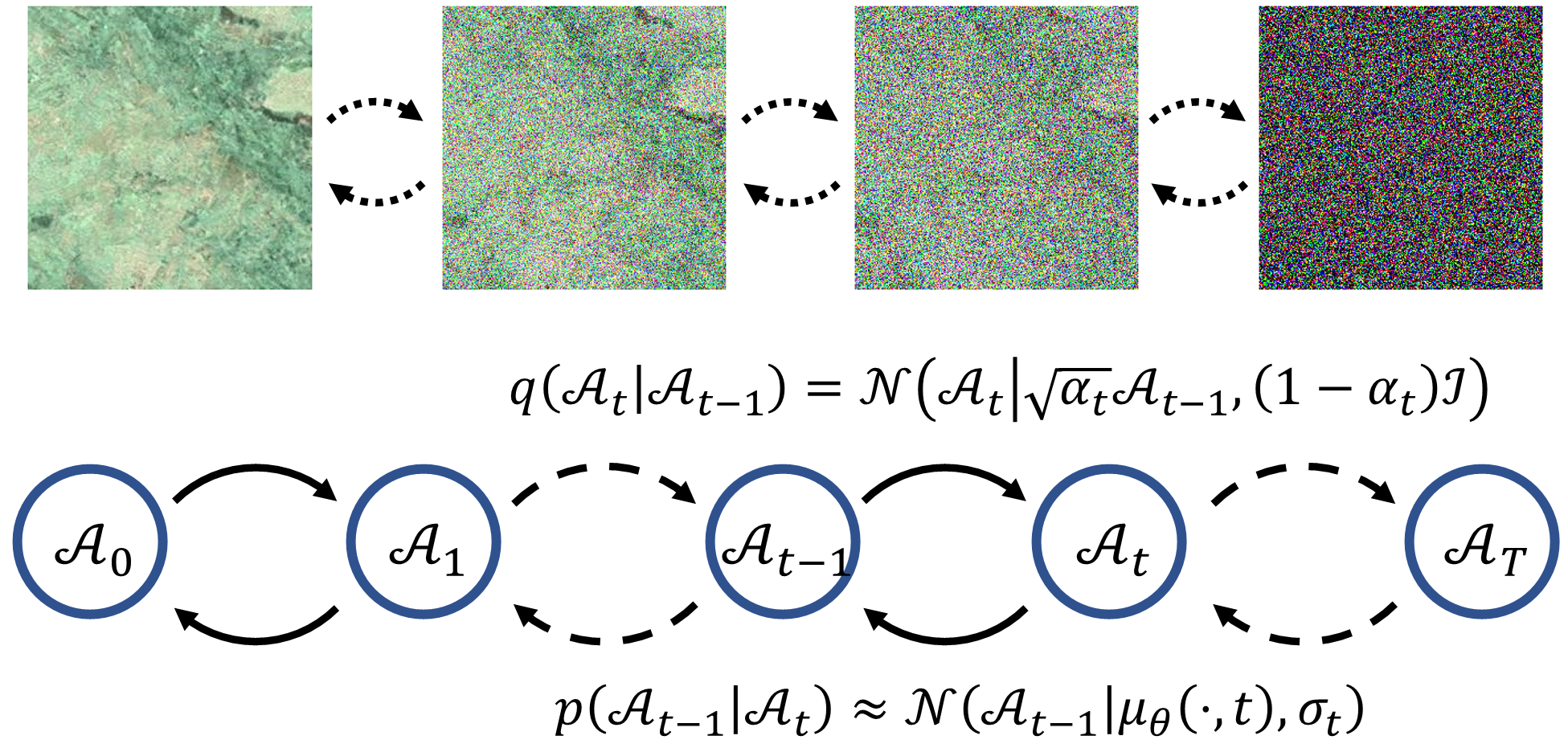}
	\caption{Denoising diffusion probabilistic model. From left to right corresponds to the forward process that gradually changes the unknown distribution $p(\A_0)$ to $p(\A_T) = \N(0,\I)$. From right to left represents the generating process that reconstructs the image.}
	\label{fig-diffu}
\end{figure}

Denoising diffusion probabilistic model (DDPM)~\cite{DDPM} is a kind of generative model. Suppose we want to sample from an unknown distribution $p(\A)$, a diffusion model introduces a Markov chain that contains a series of latent variables $\{\A_t\}_{t=1}^T$ starting from $\A_0$ (equal to $\A$ in the rest of this work) and finally reaching a variable $\A_T$ that has a known and relatively simple distribution, e.g., Gaussian distribution. The Markov chain from $\A_0$ to $\A_T$ is called the forward process in a diffusion model. The transition probability is usually assumed to be a Gaussian distribution:
\begin{align}\label{forward}
	q(\A_t|\A_{t-1}) = \N\left(\A_{t}|\sqrt{\alpha_t}\A_{t-1}, (1-\alpha_t)\I \right),~t=1,...,T,
\end{align}
where ``$\I$" is the identity tensor. The parameters $\{\alpha_t\}_{t=1}^T$ are preset as constant. ``$t$" is also called the time step and ``$T$" represents the total diffusion steps. From (\ref{forward}), one could easily find the conditional distribution about $\A_t$ and $\A_0$ as 
\begin{align}\label{forward-union}
	q(\A_t|\A_0) = \N\left(\A_{t}|\sqrt{\bar{\alpha}_t}\A_0, (1-\bar{\alpha}_t)\I\right),~t=1,...,T,
\end{align}
where $\bar{\alpha}_t := \prod_{i=1}^t\alpha_t$. If we let $\bar{\alpha}_T\approx 0$, then $\A_T$ follows a simple Gaussian distribution $\N(0,\I)$.  The forward process seems like gradually adding noise to the previous variable and finally distorting the whole data structure. The reverse of this process achieves ``generating" as we want. Specifically, starting from $\A_T$, by sampling from $p(\A_{t-1}|\A_t)$ progressively, we could reconstruct the data structure and finally reach a sample from $p(\A_0)$. However, $p(\A_{t-1}|\A_t)$ is usually intractable although the forward process is fully analytic. To solve this problem, $p(\A_{t-1}|\A_t)$ is often approximated by a Gaussian distribution 
\begin{align}\label{variational-distribution}
	p(\A_{t-1}|\A_t) = \N(\A_{t-1}|\mu_\theta(\cdot,t),\sigma_\theta(\cdot,t)).
\end{align}
In \cite{DDPM}, $\sigma_\theta(\cdot,t)$ is set as $\sqrt{1-\alpha_t}$, and $\mu_\theta(\cdot,t)$ is set as
\begin{align}\label{mu}
	\mu_\theta(\cdot,t) = \dfrac{1}{\sqrt{\alpha_t}}\left( \A_t - \dfrac{1-\alpha_t}{\sqrt{1-\bar{\alpha}_t}}\epsilon_\theta(\A_t, t) \right),  
\end{align}
so that the form of $p(\A_{t-1}|\A_t)$ matches that of $q(\A_{t-1}|\A_t,\A_0)$. ``$\epsilon_\theta(\cdot, t)$" is a neural network with parameter $\theta$. The network parameters are learned in a data-driven manner using the evidence lower bound (ELBO) as the optimization objective, which aims to minimize the distance between $p(\A_{0:T})=p(\A_T)\prod_{t=1}^Tp(\A_{t-1}|\A_t)$ and the known forward joint distribution $q(\A_{1:T}|\A_0)$:
\begin{align}\label{ELBO}
	\min_\theta~\E_{q(\A_{1:T}|\A_0)}\left[ \log\dfrac{p(\A_{0:T})}{q(\A_{1:T}|\A_0)}\right].
\end{align}
Once the training is finished, we derive an accessible parameterized form of $p(\A_{t-1}|\A_t)$ for $t=1,...,T$. Then samples from $p(\A_0)$ can be generated by ancestral sampling:
\begin{align}\label{ancestral-sampling}
	\A_{t-1} = & \dfrac{1}{\sqrt{\alpha_t}}\left( \A_t - \dfrac{1-\alpha_t}{\sqrt{1-\bar{\alpha}_t}}\epsilon_\theta(\A_t, t) \right) + \sqrt{1-\alpha_t}z_t, \nonumber\\
	& \hspace{3.2cm} z_t\sim\N(0,\I), t=T,...,1.
\end{align}
We illustrate the forward process and the generating process of the diffusion model in Fig. \ref{fig-diffu}. 

Recent work \cite{SDE} shows that as the total diffusion step ``$T$" goes infinity and the forward process becomes continuous, i.e., $\{\A(t)|t\in[0,1]\}$ indexed by continuous variable $t$, $\A(t)$ actually satisfies a stochastic differential equation (SDE):
\begin{align}
	d\A = f(\A,t)dt + g(t)d\w,
\end{align}
where $\w$ represents the standard Wiener process. For example, the diffusion process (\ref{forward}) can be seen as a discretization form of the following SDE:
\begin{align}
	d\A = -\dfrac{1}{2}(1-\alpha(t))dt + \sqrt{1-\alpha(t)}d\w.
\end{align}
The reverse of $\{\A(t)|t\in[0,1]\}$ follows the following reverse-time SDE:
\begin{align}\label{reverse-SDE}
	d\A = \left[f(\A,t) - g^2(t)\nabla_{\A(t)}\log p_t(\A(t)) \right]dt + g(t)d\bar{\w},
\end{align}
where $\bar{\w}$ is the reverse of the standard Wiener process. From the view of SDE, sampling from $p(\A(0))$ can be realized by an appropriate discretization form of (\ref{reverse-SDE}), where the ancestral sampling process (\ref{ancestral-sampling}) can be seen as a kind of such form.

\subsection{Proposed Method}
%For HS pansharpening, we propose to project the restored HRHS image $\X$ into a lower spectral dimension subspace. The corresponding coefficient $E$ can be efficiently estimated from the LRHS image $\Y$. Then we utilize a diffusion model that learns the image distribution to infer the base tensor $\A$ with the guidance of $\Y,P$ and $E$. Then the restored $\X$ is derived by projecting $\A$ back to the HS image field using $E$.

In Sec. \ref{sec-low-rank-factorization}, we have factorized the HRHS image $\X$ into two low-rank tensors. The coefficient matrix $E$ can be efficiently estimated from the LRHS image $\Y$ by (\ref{estimate-E}). Next, we will show how to infer the base tensor $\A$ by a diffusion model with the guidance of $\Y,P$ and $E$.

Take $\Y$, $P$ and estimated $E$ as conditions, we can rewrite the reverse SDE (\ref{reverse-SDE}) as 
\begin{align}\label{reverse-SDE-thiswork}
	& d\A = \left[f(\A,t) - g^2(t)\nabla_{\A(t)}\log p_t(\A(t)|\Y,P,E) \right]dt + g(t)d\bar{\w}. \nonumber \\
	& \Leftrightarrow d\A = \left\{f(\A,t) - g^2(t)[\nabla_{\A(t)}\log p_t(\A(t)) + \right. \nonumber \\
	& ~~~~~~~~~~~~~~~~ \left. \nabla_{\A(t)}\log p_t(\Y,P,E|\A(t))] \right\}dt + g(t)d\bar{\w}.
\end{align}
However, $\nabla_{\A(t)}\log p_t(\Y,P,E|\A(t))$ is intractable. Following \cite{DPS}, we approximate this term as
\begin{align}
	& \nabla_{\A(t)}\log p_t(\Y,P,E|\A(t)) \nonumber \\
	= &  \nabla_{\A(t)}\log \int p(\Y,P,E|\A(0))p(\A(0)|\A(t))d\A(0) \nonumber \\
	\approx & \nabla_{\A(t)}\log p(\Y,P,E|\hat{\A}_0),
\end{align}
where $\hat{\A}_0$ is the expectation of $\A(0)|\A(t)$ by Tweedie's formula:
\begin{align}\label{estimate-A0}
	& \hat{\A}_0(\A(t)) = \E[\A(0)|\A(t)] \nonumber\\
	& = \dfrac{1}{\sqrt{\bar{\alpha}_t}}\left[\A(t) + (1-\bar{\alpha}_t)\nabla_{\A(t)}\log p_t(\A(t))\right].
\end{align}
Combining Eq. (\ref{spectral-constraint}) and Eq. (\ref{spatial-constraint}), $\log p(\Y,P,E|\hat{\A}_0)$ can be approximated as \footnote{A precise modeling of conditional distribution $p(\Y|\X)$ should be the dirac delta distribution $\delta(\Y - \Down(\Blur(\X)))$. For computational convenience, we relax this relationship to be $\log p(\Y|\X) = -\gamma\|\Y - \Down(\Blur(\X))\|_F$. So is $p(P|\X)$.}
\begin{align}
	& \log p(\Y,P,E|\hat{\A}_0)  \nonumber\\
	= & \log p(\Y,E|\hat{\A}_0) + \log p(P|\hat{\A}_0)  \nonumber \\
	:= & -\gamma_{1} \left\|\Y\!- \Down\left(\Blur\left(\hat{\A}_0\times_3 \! E\right)\right)\right\|_F - \gamma_2 \left\|P - \hat{\A}_0\!\times_3 \! E \!\times_3 \!\sr\right\|_F \label{log-p-ype},
\end{align}
where $\gamma_1$ and $\gamma_2$ are two trade-off parameters. We discretize the reverse SDE (\ref{reverse-SDE-thiswork}) using the form of ancestral sampling process (\ref{ancestral-sampling}) as is illustrated in \cite{SDE}:
\begin{align}
	\A_{t-1} & = \dfrac{1}{\sqrt{\alpha_t}}\left( \A_t + (1-\alpha_t)\nabla_{\A(t)}\log p_t(\A(t)|\Y,P,E) \right) \nonumber \\
	& \approx \dfrac{1}{\sqrt{\alpha_t}}\left( \A_t - \dfrac{1-\alpha_t}{\sqrt{1-\bar{\alpha}_t}}\epsilon_\theta(\A_t, t) \right) +\sqrt{1-\alpha_t}z_t \nonumber\\
	& -\eta_1 \nabla_{\A_t}\left\|\Y\!- \Down\left(\Blur\left(\hat{\A}_0\times_3 \! E\right)\right)\right\|_F \nonumber \\
	& - \eta_2 \nabla_{\A_t} \left\|P - \hat{\A}_0\times_3 E \times_3 \sr \right\|_F, \label{trade-off-param}
\end{align}
where $\eta_1 = \dfrac{1-\alpha_t}{\sqrt{\alpha_t}}\gamma_1$ and $\eta_2 = \dfrac{1-\alpha_t}{\sqrt{\alpha_t}}\gamma_2$. At time $t$, we can see that the sampling consists of two parts. The first part is equal to sampling from the parameterized $p(\A_{t-1}|\A_t)$. The second part pushes the sample towards the consistent form with constraints Eq. (\ref{spectral-constraint}) and (\ref{spatial-constraint}). We summarize the proposed PLRDiff in Algorithm \ref{algo-my}. In Fig. \ref{fig-main}, we visually illustrate the flowchart of the proposed PLRDiff.

\begin{algorithm}[t]
	\renewcommand{\algorithmicrequire}{\textbf{Input:}}
	\renewcommand{\algorithmicensure}{\textbf{Initialization:}}
	\renewcommand{\algorithmicreturn}{\textbf{Output:}}
	\caption{PLRDiff for HS pansharpening}
	\label{algo-my}
	\begin{algorithmic}[1]
		\REQUIRE{LRHS image $\Y$. PAN image $P$. Index $(i_1,i_2,...,i_s)$. Trade-off parameters $\eta_1$, $\eta_2$. Total diffusion steps $T$.}
		\ENSURE{Formulate $\tilde{\A}$ and estimate the coefficient matrix $E$ by Eq. (\ref{estimate-E}). Sample $\A_T$ from $\N(0,1)$}\;
		\FOR{$t = T:1$}
		\STATE sample $\A_{t-1}$ from $p(\A_{t-1}|\A_t)$ by Eq. (\ref{variational-distribution}),
		\STATE estimate $\hat{\A}_0$ by Eq. (\ref{estimate-A0}),
		\STATE estimate $n_{t-1} = \nabla_{\A_{t-1}}\log p_{t-1}(\Y,P,E|\A_{t-1})$ using Eq. (\ref{log-p-ype})
		\STATE update $\A_{t-1} \leftarrow \A_{t-1} + \dfrac{1-\alpha_t}{\sqrt{\alpha_t}}n_{t-1}$
		\ENDFOR
		\RETURN{$\X = \A_0\times_3 E$}
	\end{algorithmic}
\end{algorithm}

\section{Experiments}\label{sec-experiments}

\begin{table}[t]
	\renewcommand{\arraystretch}{1.25}
	\newcommand{\mysize}{1.4cm}
	\fontsize{10}{11}\selectfont
	\caption{Test performance on the Chikusei dataset. The best results are in \textbf{bold}, and the second best results are with \ul{underline}.}
	\label{tab-chikusei}
	\centering
	\begin{tabular}{ M{2cm} | M{\mysize} M{\mysize} M{\mysize} M{\mysize} M{\mysize} M{\mysize}}
		\Xhline{0.8pt}
        methods & PSNR$\uparrow$ & SSIM$\uparrow$ & SAM$\downarrow$ & ERGAS$\downarrow$ & SCC$\uparrow$ & Q2N$\uparrow$ \\
        \hline 
		CNMF     & \ul{29.65} & 0.7897 & \tb{4.1230} & \ul{4.7869} & \ul{0.7608} & 0.5464\\
		HySure   & 29.16 & 0.7883 & 4.4721 & 5.0951 & 0.7358 & \ul{0.5548}\\
		GLP      & 29.08 & 0.7892 & 4.5450 & 5.0006 & 0.6887 & 0.5250\\
		Brovey   & 26.47 & 0.7367 & \ul{4.1615} & 6.3723 & 0.6938 & 0.2055\\
		GS       & 25.56 & 0.6967 & 5.1679 & 7.3331 & 0.6169 & 0.1839\\
		AWLP     & 28.40 & 0.7782 & 5.6753 & 5.4027 & 0.6252 & 0.5083\\
		SFIM     & 23.37 & \ul{0.7924} & 4.7497 & 312.14 & 0.0475 & 0.4976\\
		FUSE     & 25.28 & 0.6373 & 4.8927 & 8.4065 & 0.5611 & 0.3603\\
		CNNFUS   & 26.93 & 0.7072 & 4.8100 & 6.3454 & 0.6175 & 0.4998\\
		PLRDiff     & \tb{30.81} & \tb{0.8184} & 4.2079 & \tb{4.6715} & \tb{0.8133} & \tb{0.6170}\\
		\Xhline{0.8pt}
	\end{tabular}
\end{table}

\begin{table}[t]
	\renewcommand{\arraystretch}{1.25}
	\newcommand{\mysize}{1.4cm}
	\fontsize{10}{11}\selectfont
	\caption{Test performance on the Houston dataset. The best results are in \textbf{bold}, and the second best results are with \ul{underline}.}
	\label{tab-houston}
	\centering
	\begin{tabular}{ M{2cm} | M{\mysize} M{\mysize} M{\mysize} M{\mysize} M{\mysize} M{\mysize}}
		\Xhline{0.8pt}
		methods & PSNR$\uparrow$ & SSIM$\uparrow$ & SAM$\downarrow$ & ERGAS$\downarrow$ & SCC$\uparrow$ & Q2N$\uparrow$ \\
		\hline 
		CNMF   & 35.47 & 0.8831 & \ul{4.9209} & 4.1782 & 0.9067 & 0.7473 \\
		HySure   & 34.96 & 0.8711 & 6.1476 & 4.4256 & 0.8977 & 0.8482\\
		GLP   & \ul{36.57} & \ul{0.8982} & 5.0029 & \tb{3.6505} & \ul{0.9133} & 0.8534\\
		Brovey   & 33.17 & 0.8604 & 5.9299 & 5.3222 & 0.8741 & 0.7759\\
		GS   & 33.73 & 0.8801 & 5.0933 & 4.9603 & 0.9066 & 0.8001\\
		AWLP   & 35.59 & 0.8896 & 5.5949 & 4.0023 & 0.8963 & 0.6888\\
		SFIM   & 34.92 & 0.8841 & 5.4133 & 4.1640 & 0.8971 & 0.8026\\
		FUSE   & 29.74 & 0.7293 & 6.7537 & 8.2079 & 0.5376 & 0.6094\\
		CNNFUS   & 31.97 & 0.8149 & 6.3689 & 5.8997 & 0.8105 & 0.7568\\
		PLRDiff   & \tb{37.76} & \tb{0.9037} & \tb{4.8759} & \ul{3.9868} & \tb{0.9160} & \tb{0.9111}\\
		\Xhline{0.8pt}
	\end{tabular}
\end{table}

\begin{table}[t]
	\renewcommand{\arraystretch}{1.25}
	\newcommand{\mysize}{1.4cm}
	\fontsize{10}{11}\selectfont
	\caption{Test performance on the Pavia dataset. The best results are in \textbf{bold}, and the second best results are with \ul{underline}.}
	\label{tab-pavia}
	\centering
	\begin{tabular}{ M{2cm} | M{\mysize} M{\mysize} M{\mysize} M{\mysize} M{\mysize} M{\mysize}}
		\Xhline{0.8pt}
		methods & PSNR$\uparrow$ & SSIM$\uparrow$ & SAM$\downarrow$ & ERGAS$\downarrow$ & SCC$\uparrow$ & Q2N$\uparrow$ \\
		\hline 
		CNMF   & 29.08 & 0.8500 & 6.6030 & 6.4761 & \ul{0.7481} & 0.2488\\
		HySure   & 27.96 & 0.7992 & 7.7375 & 7.4845 & 0.6699 & 0.3618\\
		GLP   & 28.29 & 0.8405 & 7.7576 & 7.1214 & 0.7214 & 0.3273\\
		Brovey   & 26.58 & 0.7536 & 7.1325 & 8.5029 & 0.7103 & 0.3024\\
		GS   & 26.56 & 0.7379 & 7.7288 & 8.4432 & 0.7341 & 0.3238\\
		AWLP   & 27.72 & 0.8237 & 8.7682 & 7.5364 & 0.6811 & 0.2982\\
		SFIM   & 27.77 & 0.8069 & 6.9379 & 7.3315 & 0.7108 & 0.3204\\
		FUSE   & 28.05 & 0.8301 & \ul{6.0479} & 7.1051 & 0.6528 & 0.2299\\
		CNNFUS   & \ul{30.37} & \ul{0.9075} & \tb{5.9648} & \ul{5.6894} & 0.7267 & \ul{0.4030}\\
		PLRDiff   & \tb{32.39} & \tb{0.9401} & 6.4635 & \tb{4.5579} & \tb{0.8402} & \tb{0.4624}\\
		\Xhline{0.8pt}
	\end{tabular}
\end{table}

In this section, we conduct a series of experiments to verify the effectiveness and superiority of the proposed PLRDiff method. For PLRDiff, we utilize the pre-trained diffusion model\footnote{Pre-trained diffusion model is available at \url{https://github.com/wgcban/ddpm-cd}.}. The diffusion model is trained using remote-sensing images processed by the Google Earth Engine. The spectral dimensionality of the generated image is three. We use three widely used datasets for testing, namely:

(1) The Chikusei dataset\footnote{\href{https://naotoyokoya.com/Download.html}{https://naotoyokoya.com/Download.html}} contains a HS image of size $2517\times 2335\times 128$. It was taken by a Headwall Hyperspec-VNIR-C imaging sensor with a spectral range from 363nm to 1018nm. We crop the centre part and derive the HRHS image of size $256\times 256 \times 128$. The LRHS and PAN images are generated by Wald’s protocol \cite{Wald1}\cite{Wald2}. Specifically, the LRHS image is generated by spatially blurring the HRHS image using a Gaussian filter of size $9\times 9$ and downsampling the processed result with a scale of 4. We average the visible bands of the HRHS image to generate the PAN image.  

(2) The Houston dataset\footnote{\href{https://hyperspectral.ee.uh.edu/?page_id=459}{https://hyperspectral.ee.uh.edu/?page\_id=459}} was acquired by the NSF-funded Center for Airborne Laser Mapping (NCALM). The spatial size of the HS image is $349\times 1905$ at a resolution of 2.5m. Its spectrum ranges from 380nm to 1050nm. The centre part with the size of $256\times256\times 144$ is taken as an HRHS image. We generate the LRHS and PAN images using the same method as for the Chikusei dataset. 

(3) The Pavia Center dataset was captured by Reflective Optics System Imaging Spectrometer (ROSIS). The spectrum ranges from 400nm to 900 nm. We use the data provided by the website \footnote{\href{https://github.com/liangjiandeng/HyperPanCollection}{https://github.com/liangjiandeng/HyperPanCollection}}. There are two sets of data in this dataset. Each contains an HRHS image of size $400\times 400\times 102$, a LRHS image of size $100\times 100\times 102$ and a PAN image of size $400\times 400$.

%\footnote{\href{https://www.ehu.eus/ccwintco/index.php/Hyperspectral_Remote_Sensing_Scenes}{https://www.ehu.eus/ccwintco/index.php/Hyperspectral\_Remote\_Sensing \\ \_Scenes}}

Six quantitative indices are used to evaluate the restoration results. The Peak Signal-to-Noise Ratio (PSNR) assesses perceptual image quality. The Structural Similarity Index Measure (SSIM) compares the structural information between two images. The Q2N \cite{Q2N} index is particularly designed for measuring quality for multiband images. The spectral Angle Mapper (SAM) measures the spectral similarity between two HS images. The Error Relative Global Dimension Synthesis (ERGAS) evaluates the quality of the HS image by
a normalized average error of each band. The spatial correlation coefficient (SCC) measures the spatial details preserved in the restored image.

To estimate the blur operator $\Blur$ and spectral response vector $\sr$, we parameterized them by $\Blur_\theta = \mathrm{softmax}(\theta)$ and $\sr_\eta = \mathrm{softmax}(\eta)$. $\theta\in\R^{k\times k}$ and $\eta\in\R^{1\times s}$ are parameters, where $k$ represents the blur kernel size. Then, $\theta$ and $\eta$ are optimized by 
\begin{align}
	\min_{\theta, \eta}~\left\| \Down(\Blur_{\theta}(P)) - \Y\times_3 \sr_{\eta}\right\|_F^2.
\end{align}

\begin{figure}[ht]
	\newcommand{\mysize}{6cm}
	\centering
	\begin{minipage}[t]{\mysize}
		\centering
		\fontsize{10}{11}\selectfont
		\includegraphics[width=\mysize]{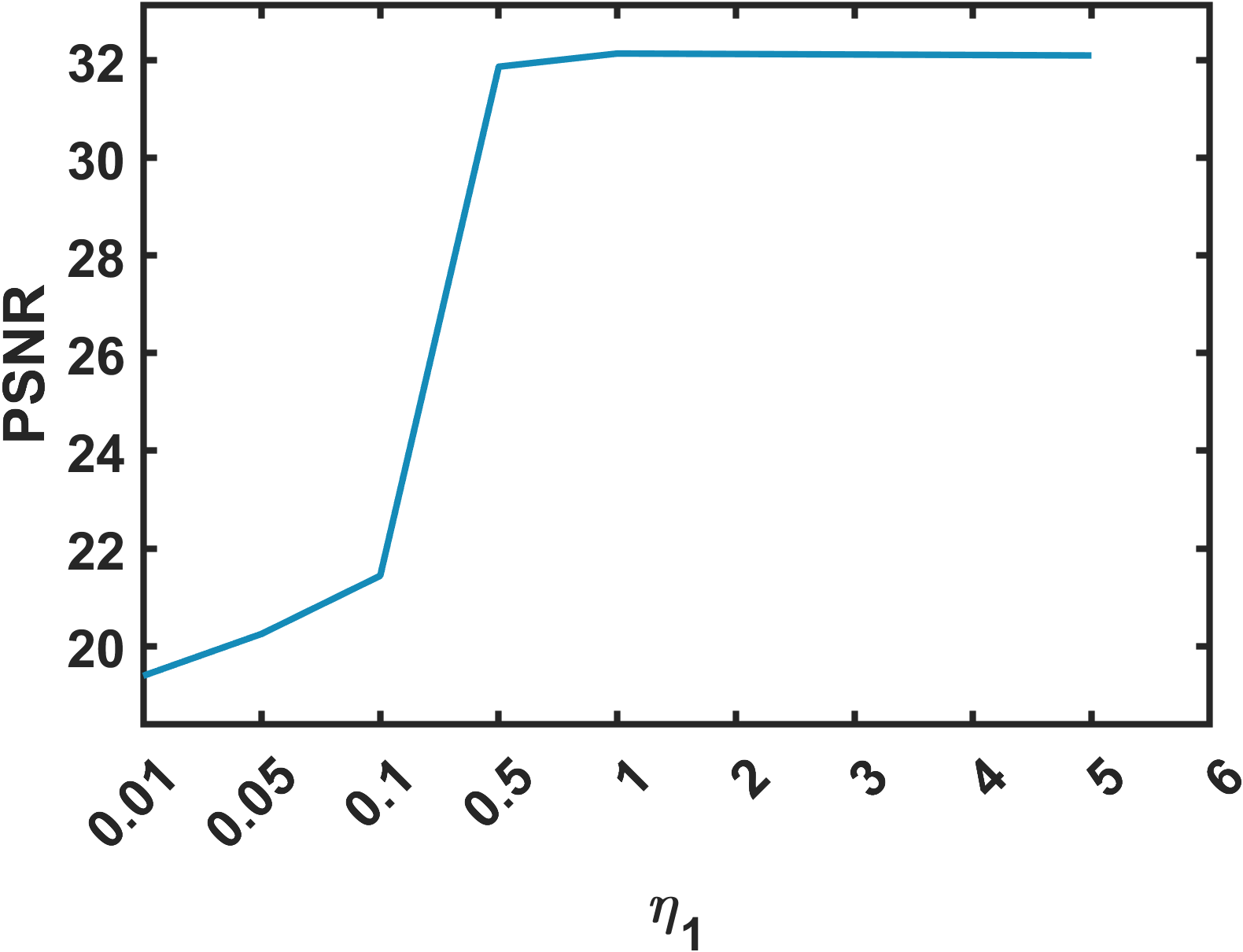} 
	\end{minipage} \hspace{4pt}
	\begin{minipage}[t]{\mysize}
		\centering
		\fontsize{10}{11}\selectfont
		\includegraphics[width=\mysize]{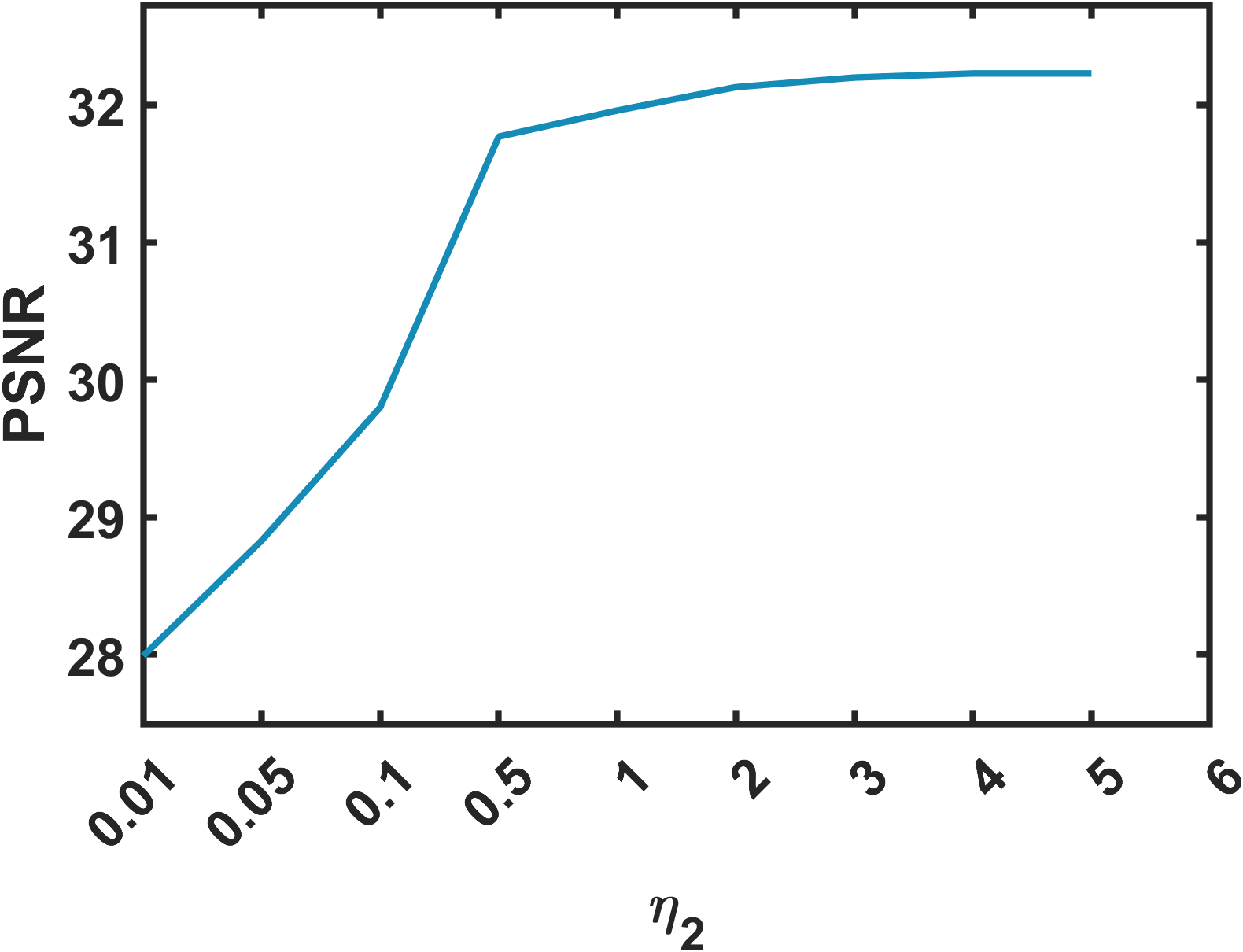} 
	\end{minipage} 
	\caption{Sensitivity analysis of the parameters $\eta_1$ and $\eta_2$.}
	\label{fig-eta}
\end{figure}

The proposed PLRDiff contains two trade-off parameters $\eta_1$ and $\eta_2$. We validate the parameters on the Chikusei validation dataset. The validation dataset is generated using the same way as for the testing dataset, while the corresponding HRMS image is cropped from other part of the Chikusei dataset. The sensitivity of $\eta_1$ and $\eta_2$ is analyzed in Fig. \ref{fig-eta}. For all the experiments, we set them as $\eta_1 = 2$ and $\eta_2 = 2$. The total diffusion step $T$ is set as $500$.

\begin{figure}[ht]
	\newcommand{\mysize}{3cm}
	\centering
	\begin{minipage}[t]{\mysize}
		\centering
		\fontsize{10}{11}\selectfont
		\includegraphics[width=\mysize]{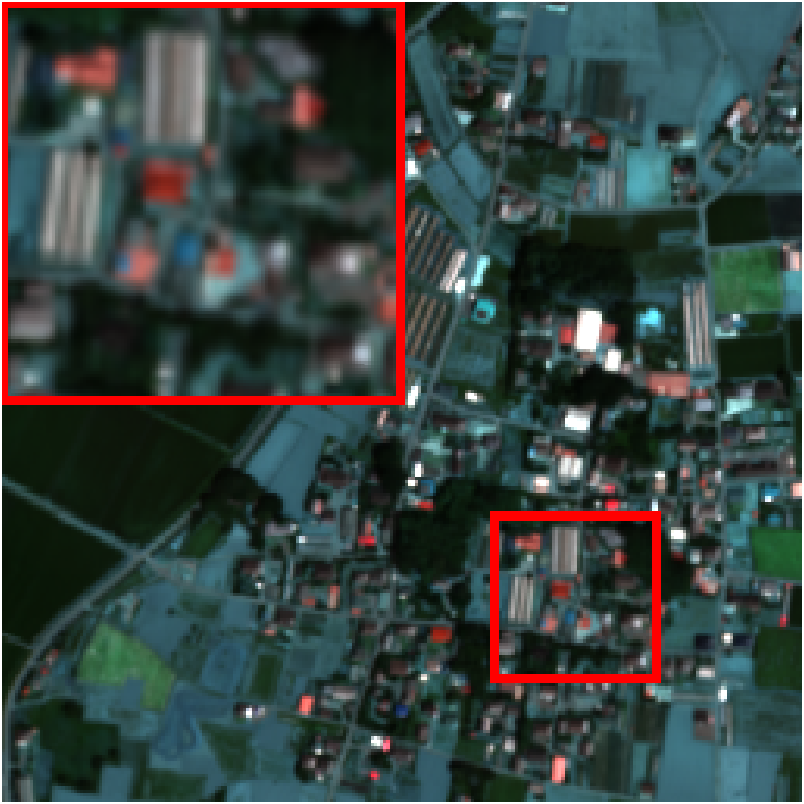} \\
		HRHS
	\end{minipage}
	\begin{minipage}[t]{\mysize}
		\centering
		\fontsize{10}{11}\selectfont
		\includegraphics[width=\mysize]{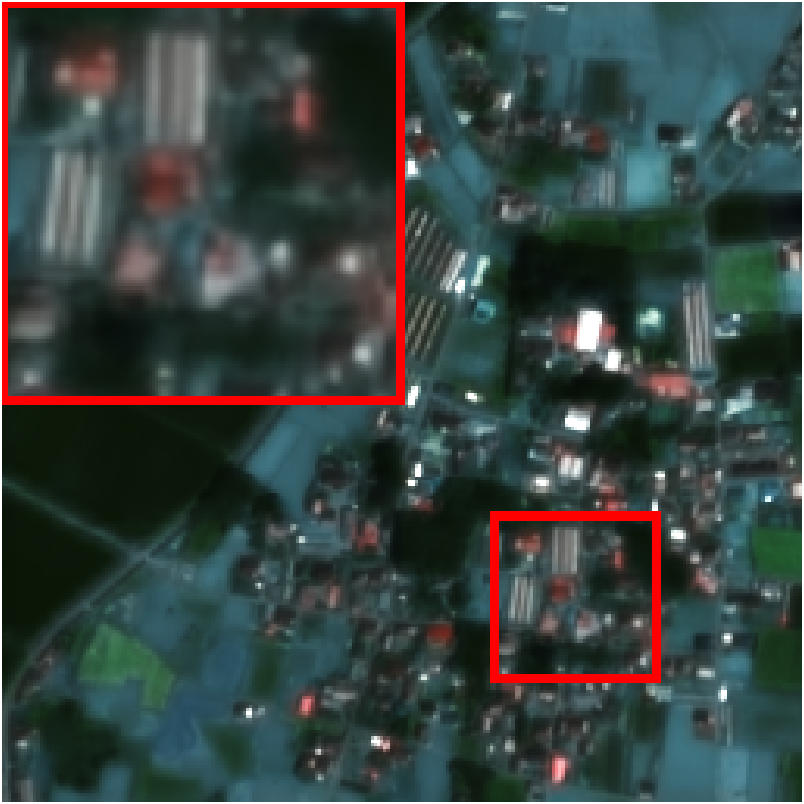} \\
		CNMF
	\end{minipage}
	\begin{minipage}[t]{\mysize}
		\centering
		\fontsize{10}{11}\selectfont
		\includegraphics[width=\mysize]{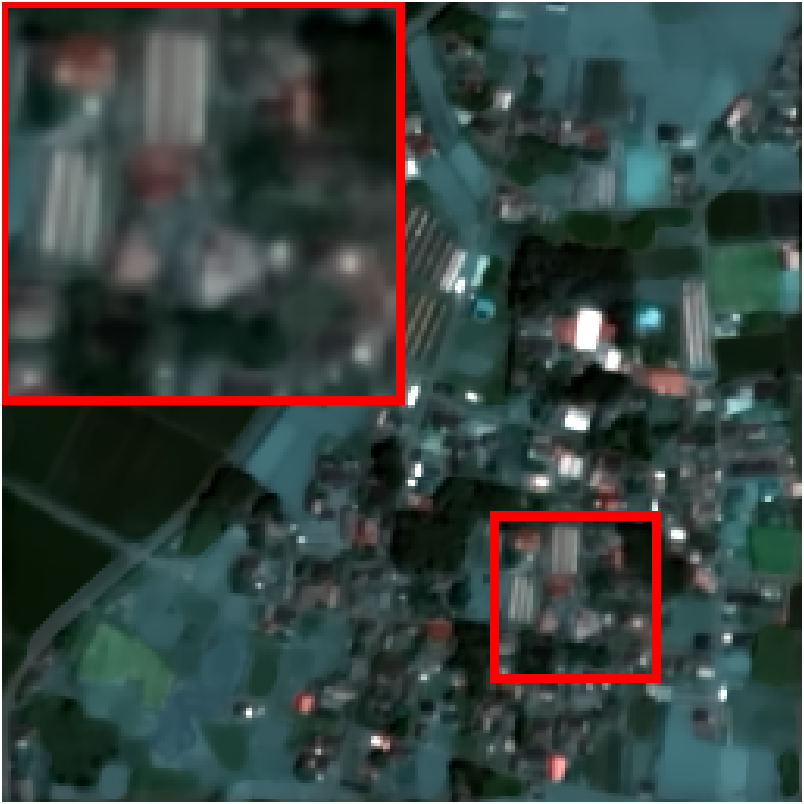} \\
		HySure
	\end{minipage}
	\begin{minipage}[t]{\mysize}
		\centering
		\fontsize{10}{11}\selectfont
		\includegraphics[width=\mysize]{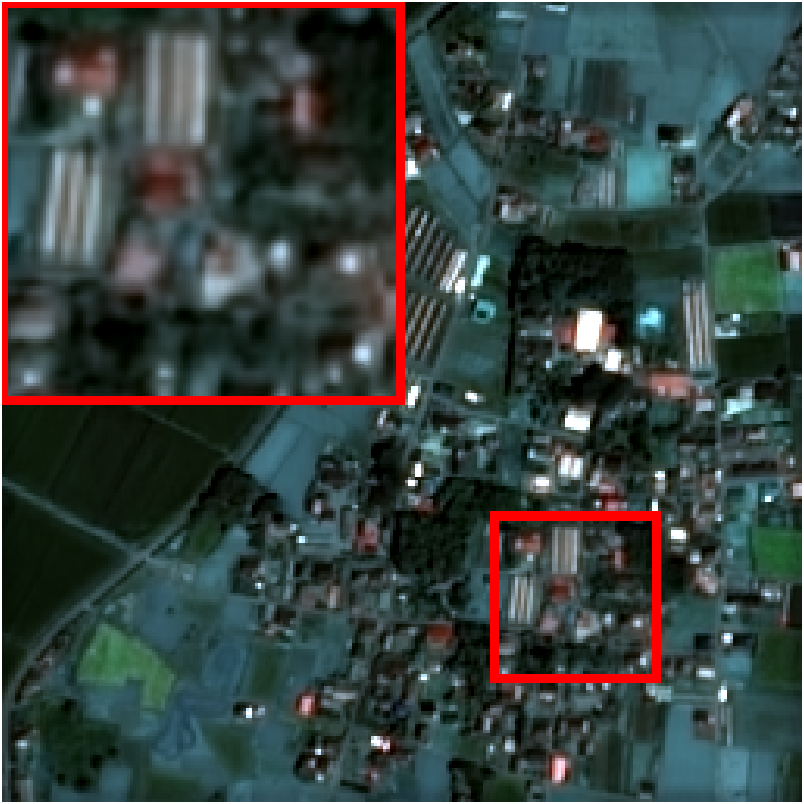} \\
		GLP
	\end{minipage} \vspace{3pt} \\

	\begin{minipage}[t]{\mysize}
		\centering
		\fontsize{10}{11}\selectfont
		\includegraphics[width=\mysize]{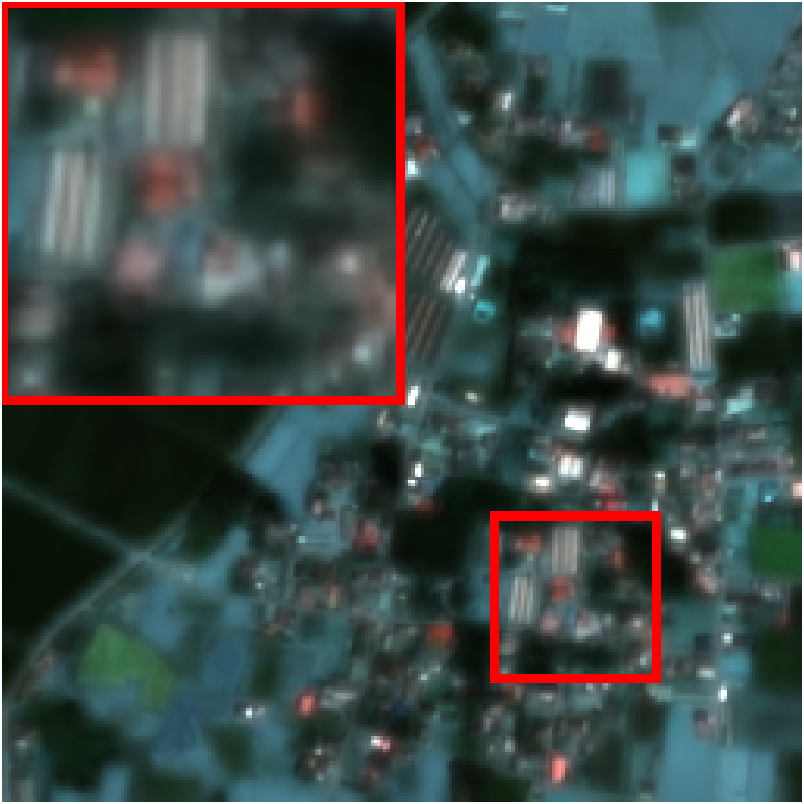} \\
		Brovey
	\end{minipage}
	\begin{minipage}[t]{\mysize}
		\centering
		\fontsize{10}{11}\selectfont
		\includegraphics[width=\mysize]{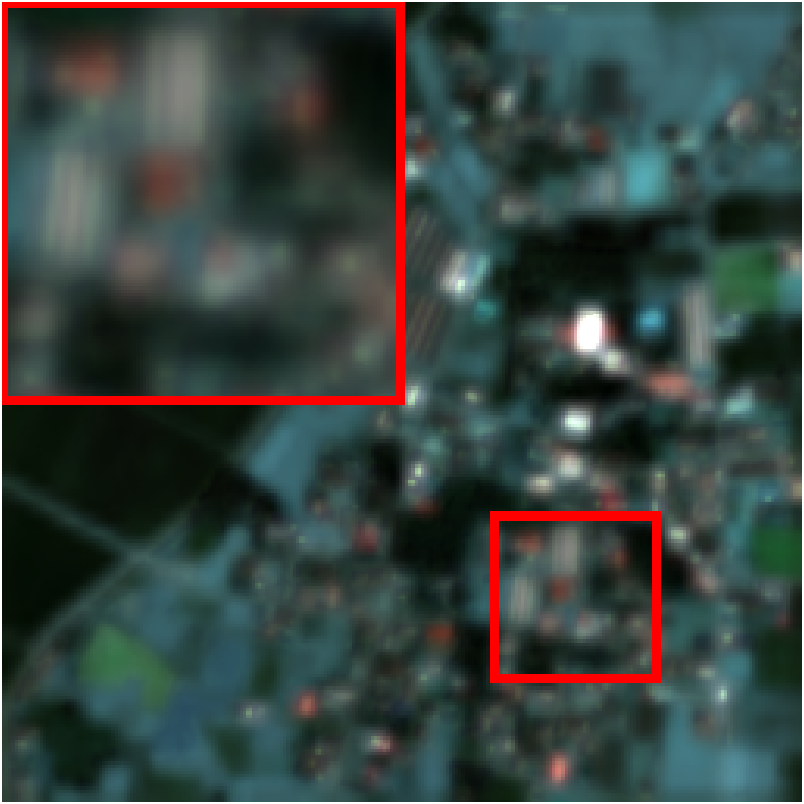} \\
		GS
	\end{minipage} 
	\begin{minipage}[t]{\mysize}
		\centering
		\fontsize{10}{11}\selectfont
		\includegraphics[width=\mysize]{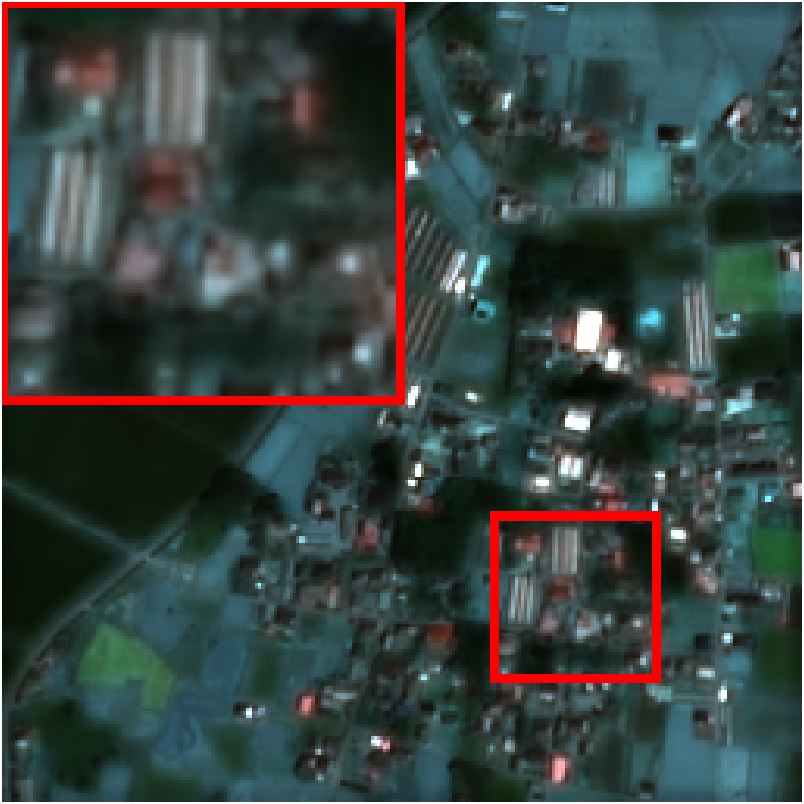} \\
		AWLP
	\end{minipage}
	\begin{minipage}[t]{\mysize}
		\centering
		\fontsize{10}{11}\selectfont
		\includegraphics[width=\mysize]{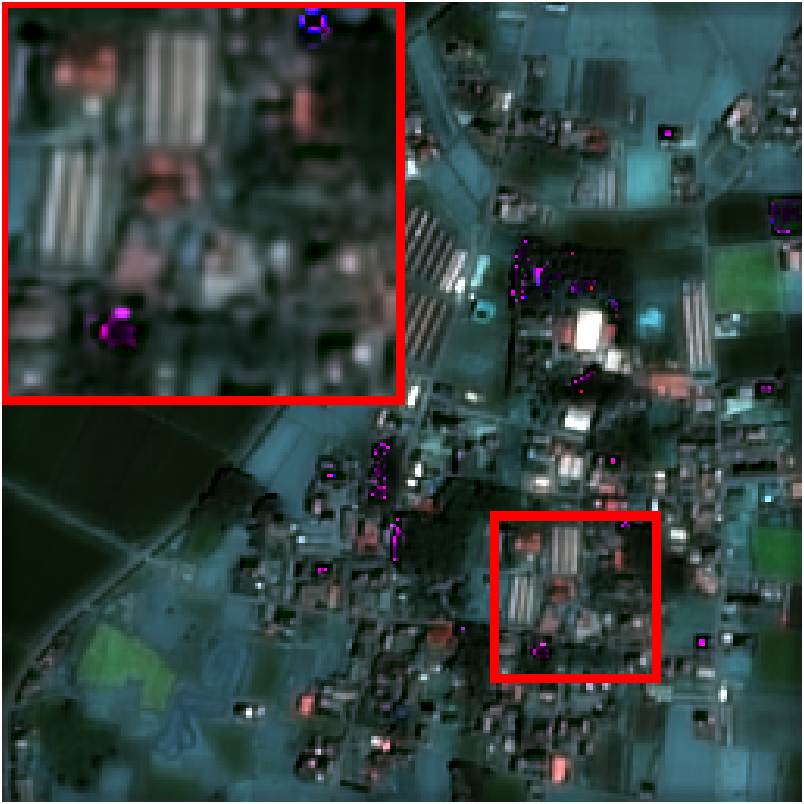} \\
		SFIM 
	\end{minipage}  \vspace{3pt} \\
	\begin{minipage}[t]{\mysize}
		\centering
		\fontsize{10}{11}\selectfont
		\includegraphics[width=\mysize]{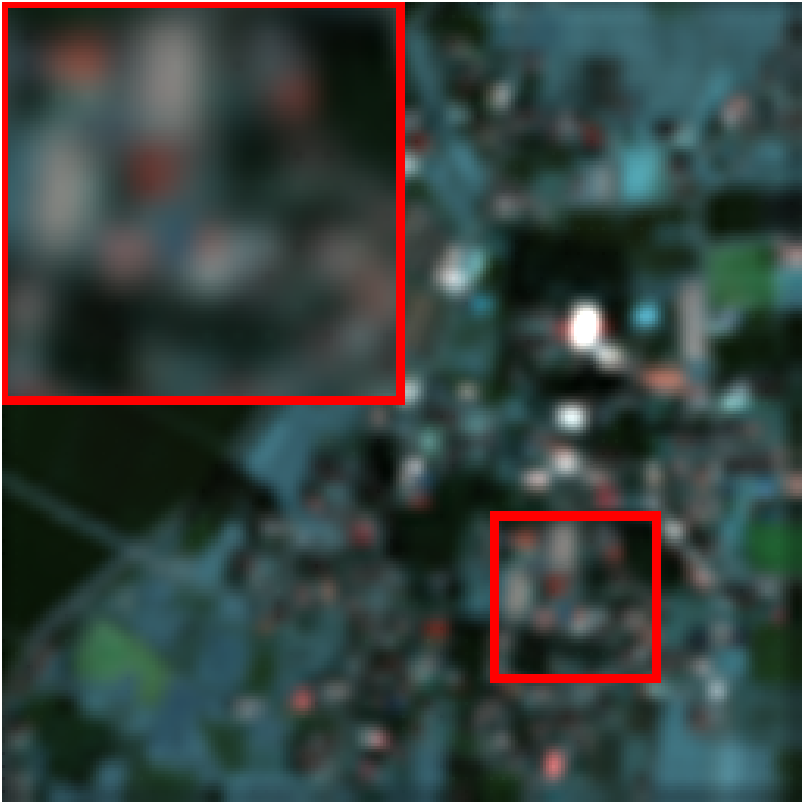} \\
		FUSE
	\end{minipage}
	\begin{minipage}[t]{\mysize}
		\centering
		\fontsize{10}{11}\selectfont
		\includegraphics[width=\mysize]{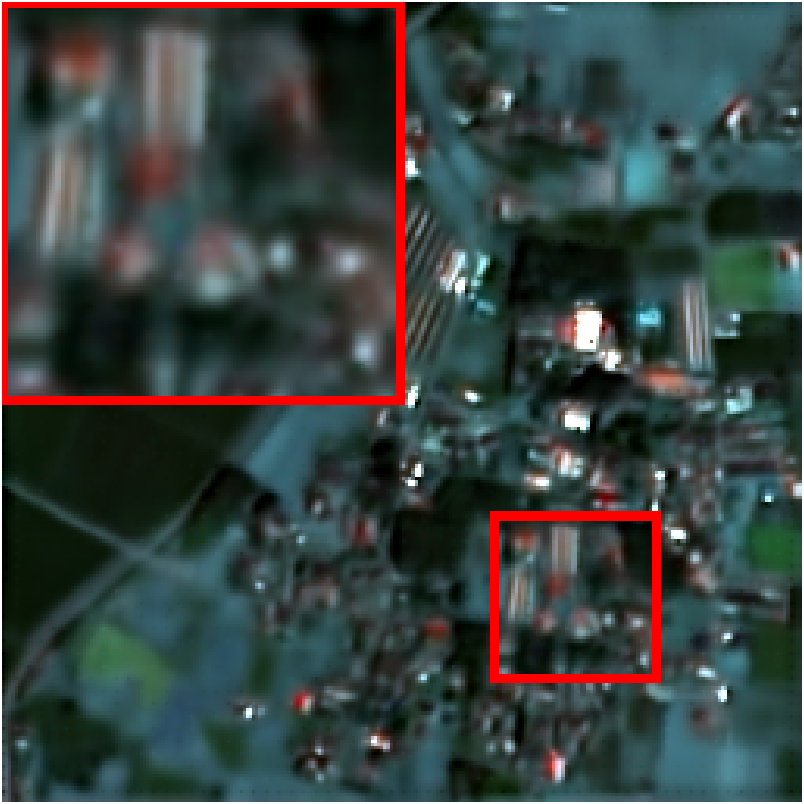} \\
		CNNFUS
	\end{minipage}
	\begin{minipage}[t]{\mysize}
		\centering
		\fontsize{10}{11}\selectfont
		\includegraphics[width=\mysize]{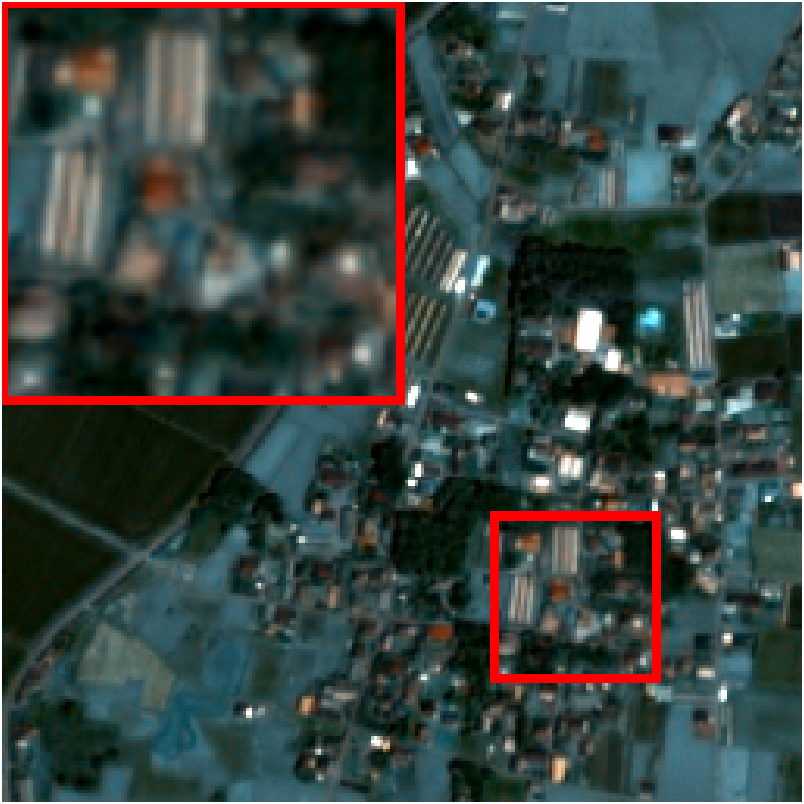} \\
		Ours
	\end{minipage}
	\caption{Visual results on Chikusei dataset. Shown bands are [20,50,60].}
	\label{fig-chi}
\end{figure}

\begin{figure}[t]
	\newcommand{\mysize}{3cm}
	\centering
	\begin{minipage}[t]{\mysize}
		\centering
		\fontsize{10}{11}\selectfont
		\includegraphics[width=\mysize]{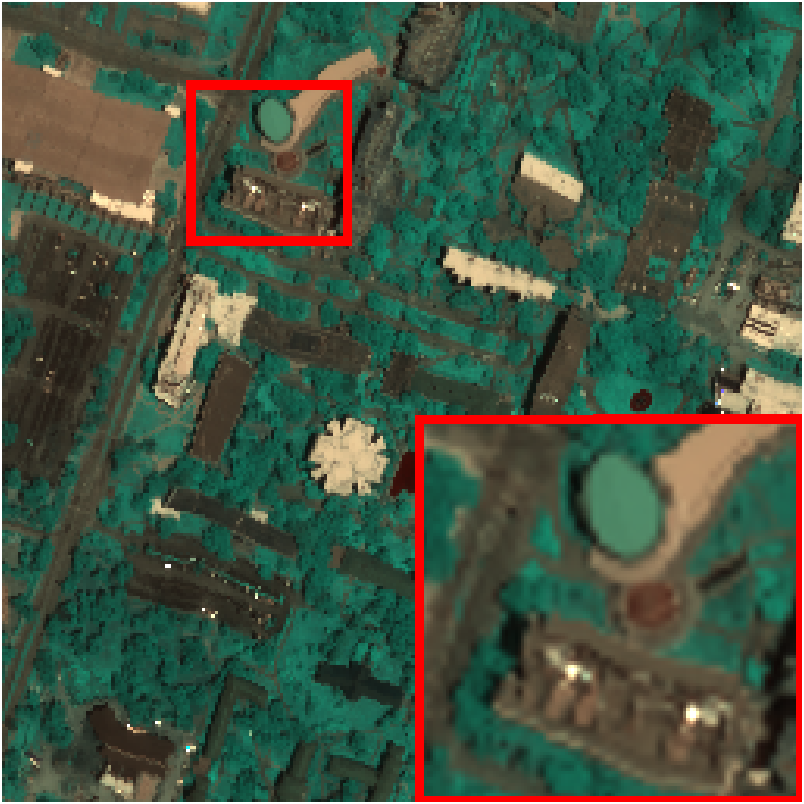} \\
		HRHS
	\end{minipage}
	\begin{minipage}[t]{\mysize}
		\centering
		\fontsize{10}{11}\selectfont
		\includegraphics[width=\mysize]{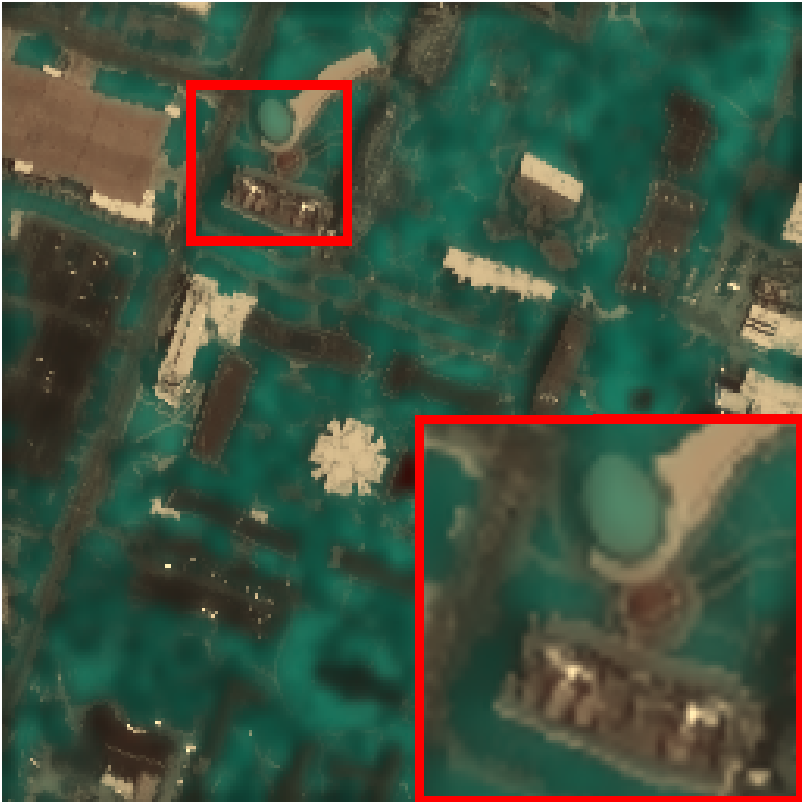} \\
		CNMF
	\end{minipage}
	\begin{minipage}[t]{\mysize}
		\centering
		\fontsize{10}{11}\selectfont
		\includegraphics[width=\mysize]{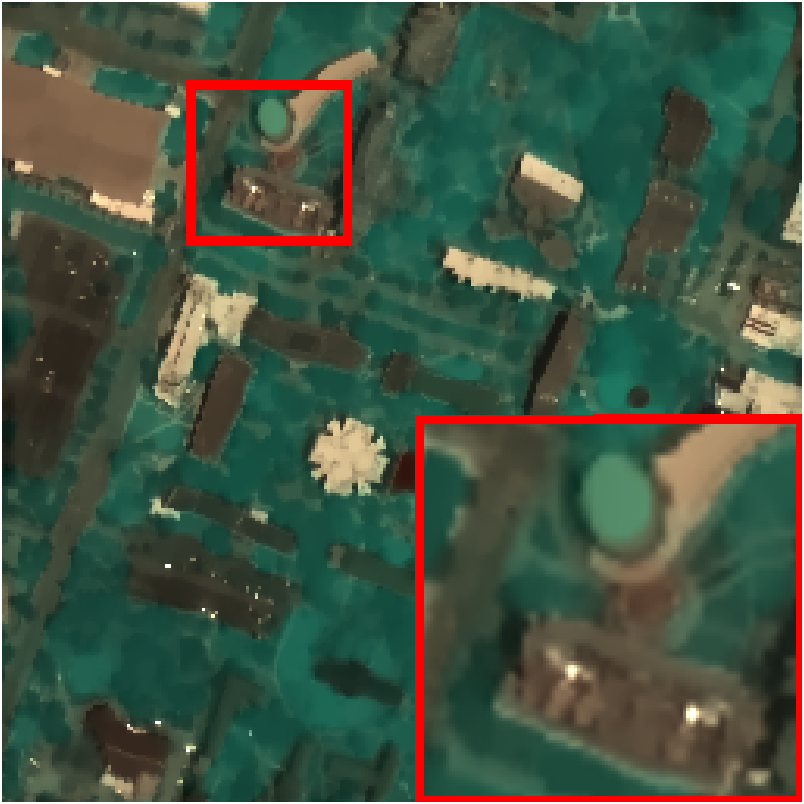} \\
		HySure
	\end{minipage}
	\begin{minipage}[t]{\mysize}
		\centering
		\fontsize{10}{11}\selectfont
		\includegraphics[width=\mysize]{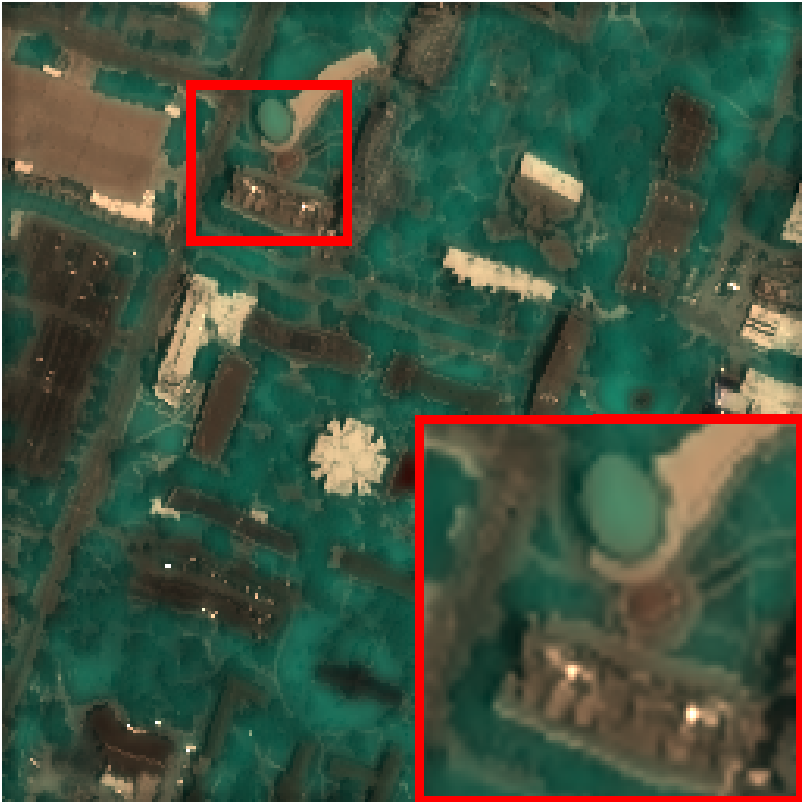} \\
		GLP
	\end{minipage} \vspace{3pt} \\

	\begin{minipage}[t]{\mysize}
		\centering
		\fontsize{10}{11}\selectfont
		\includegraphics[width=\mysize]{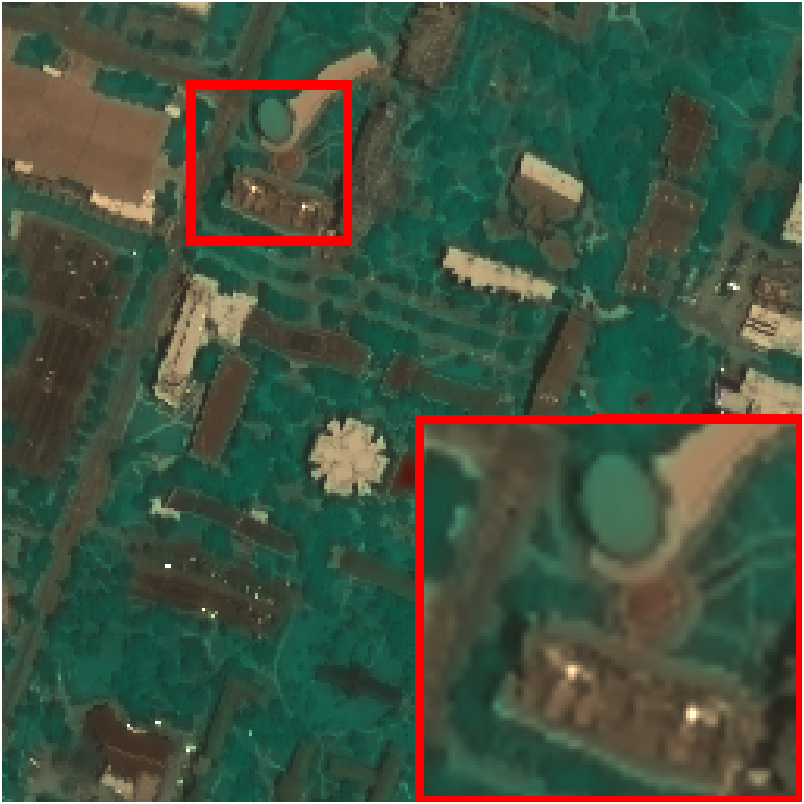} \\
		Brovey
	\end{minipage}
	\begin{minipage}[t]{\mysize}
		\centering
		\fontsize{10}{11}\selectfont
		\includegraphics[width=\mysize]{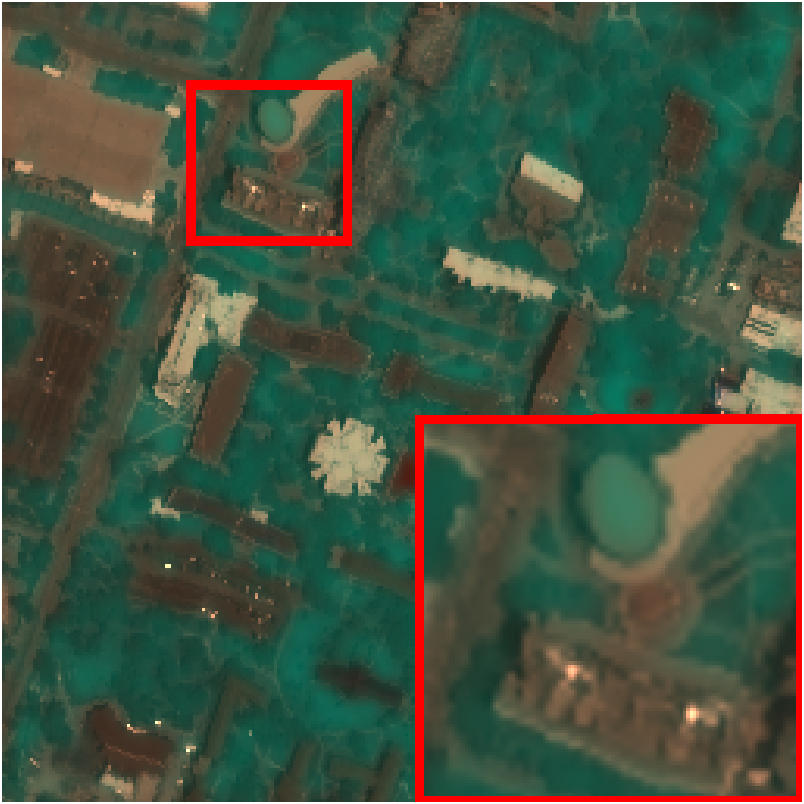} \\
		GS
	\end{minipage}	
	\begin{minipage}[t]{\mysize}
		\centering
		\fontsize{10}{11}\selectfont
		\includegraphics[width=\mysize]{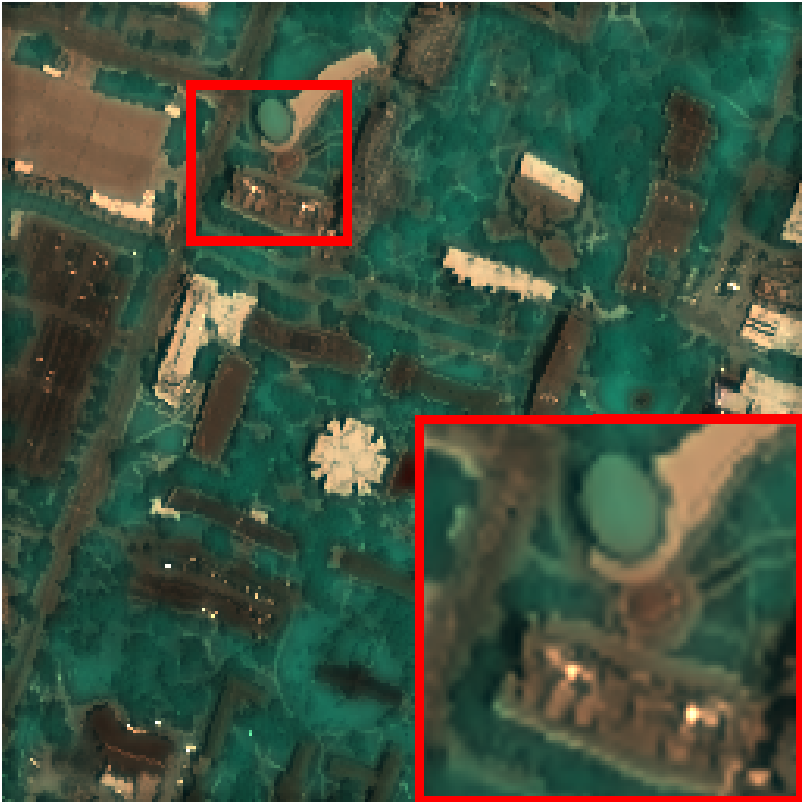} \\
		AWLP
	\end{minipage}
	\begin{minipage}[t]{\mysize}
		\centering
		\fontsize{10}{11}\selectfont
		\includegraphics[width=\mysize]{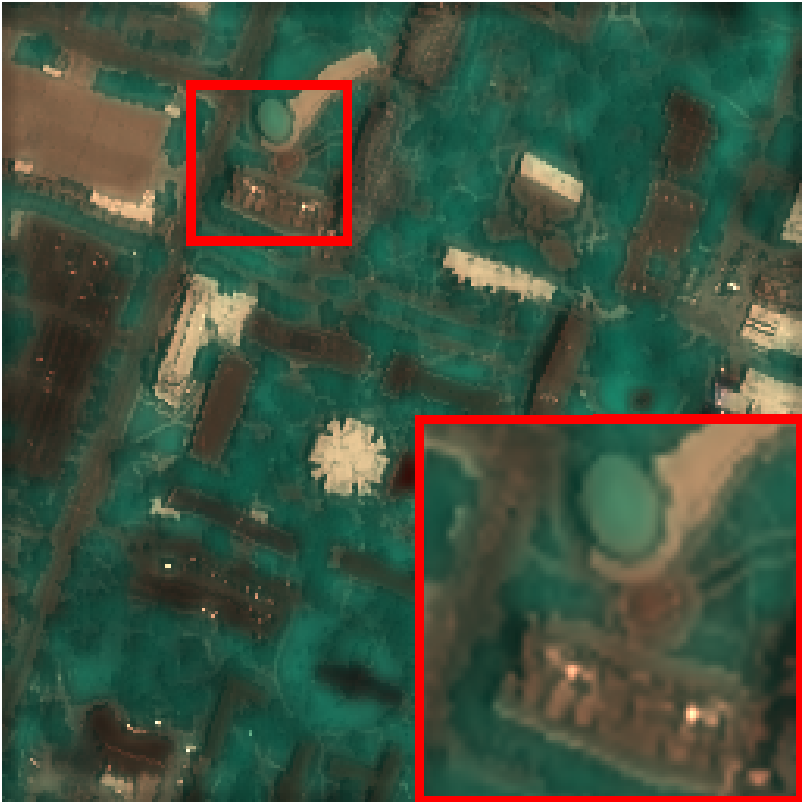} \\
		SFIM 
	\end{minipage} \vspace{3pt} \\
	\begin{minipage}[t]{\mysize}
		\centering
		\fontsize{10}{11}\selectfont
		\includegraphics[width=\mysize]{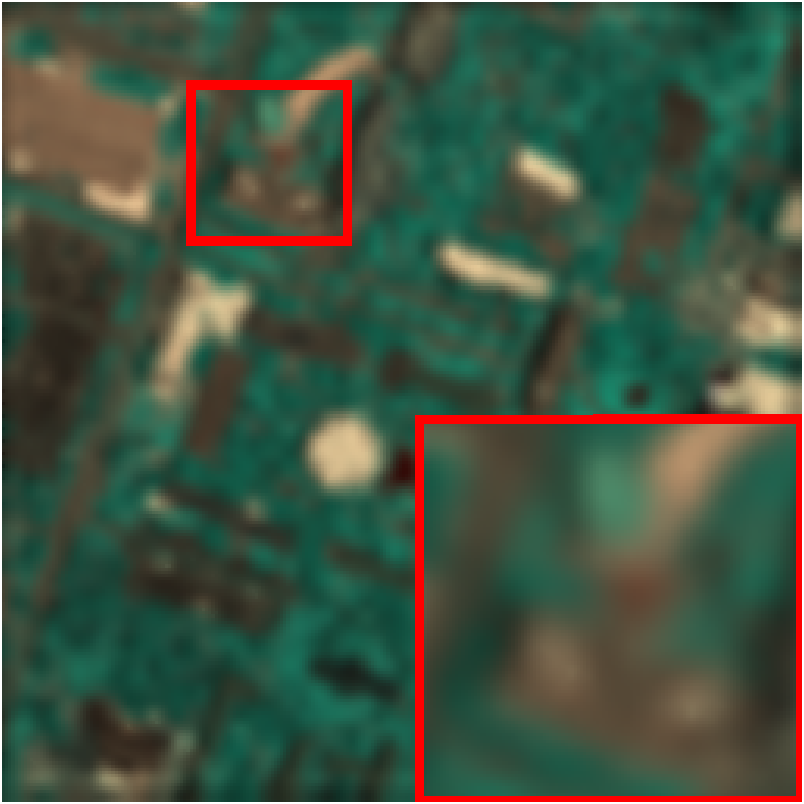} \\
		FUSE
	\end{minipage}
	\begin{minipage}[t]{\mysize}
		\centering
		\fontsize{10}{11}\selectfont
		\includegraphics[width=\mysize]{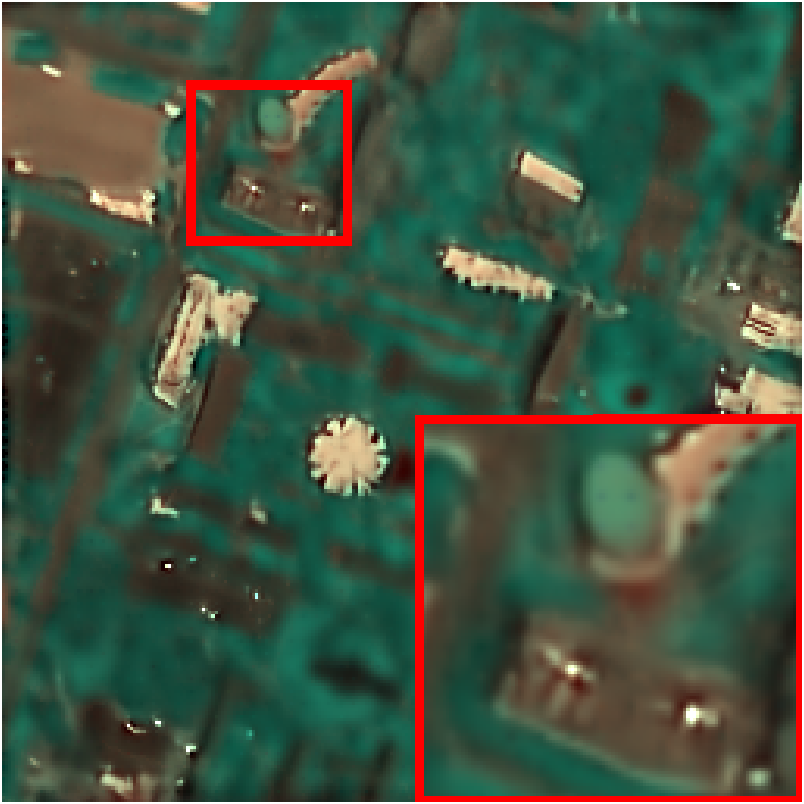} \\
		CNNFUS
	\end{minipage}
	\begin{minipage}[t]{\mysize}
		\centering
		\fontsize{10}{11}\selectfont
		\includegraphics[width=\mysize]{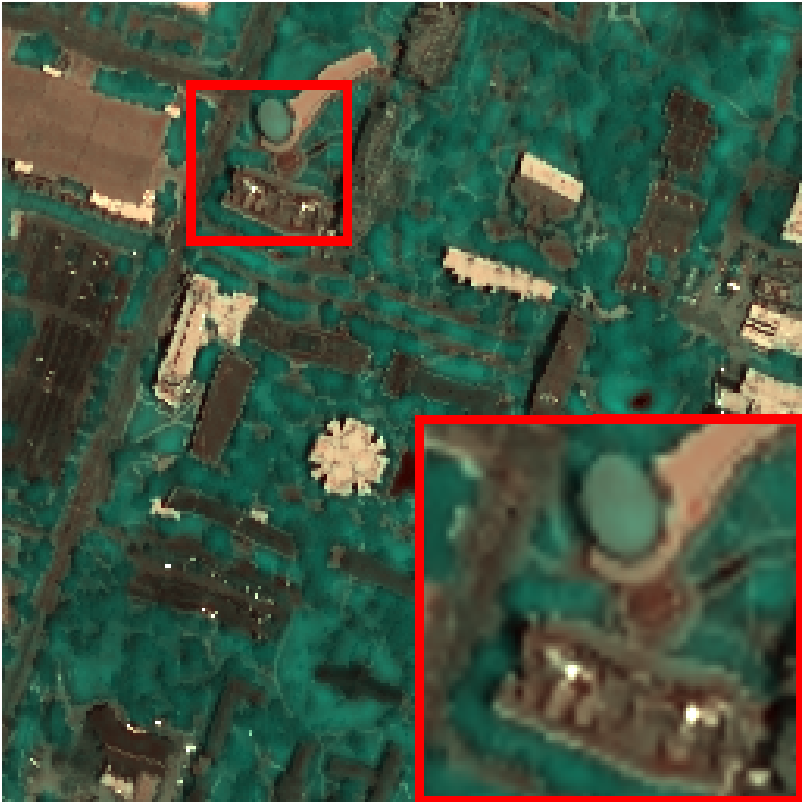} \\
		Ours
	\end{minipage}
	\caption{Visual results on Houston dataset. Shown bands are [50,90,100].}
	\label{fig-hou}
\end{figure}

\begin{figure}[t]
	\newcommand{\mysize}{3cm}
	\centering
	\begin{minipage}[t]{\mysize}
		\centering
		\fontsize{10}{11}\selectfont
		\includegraphics[width=\mysize]{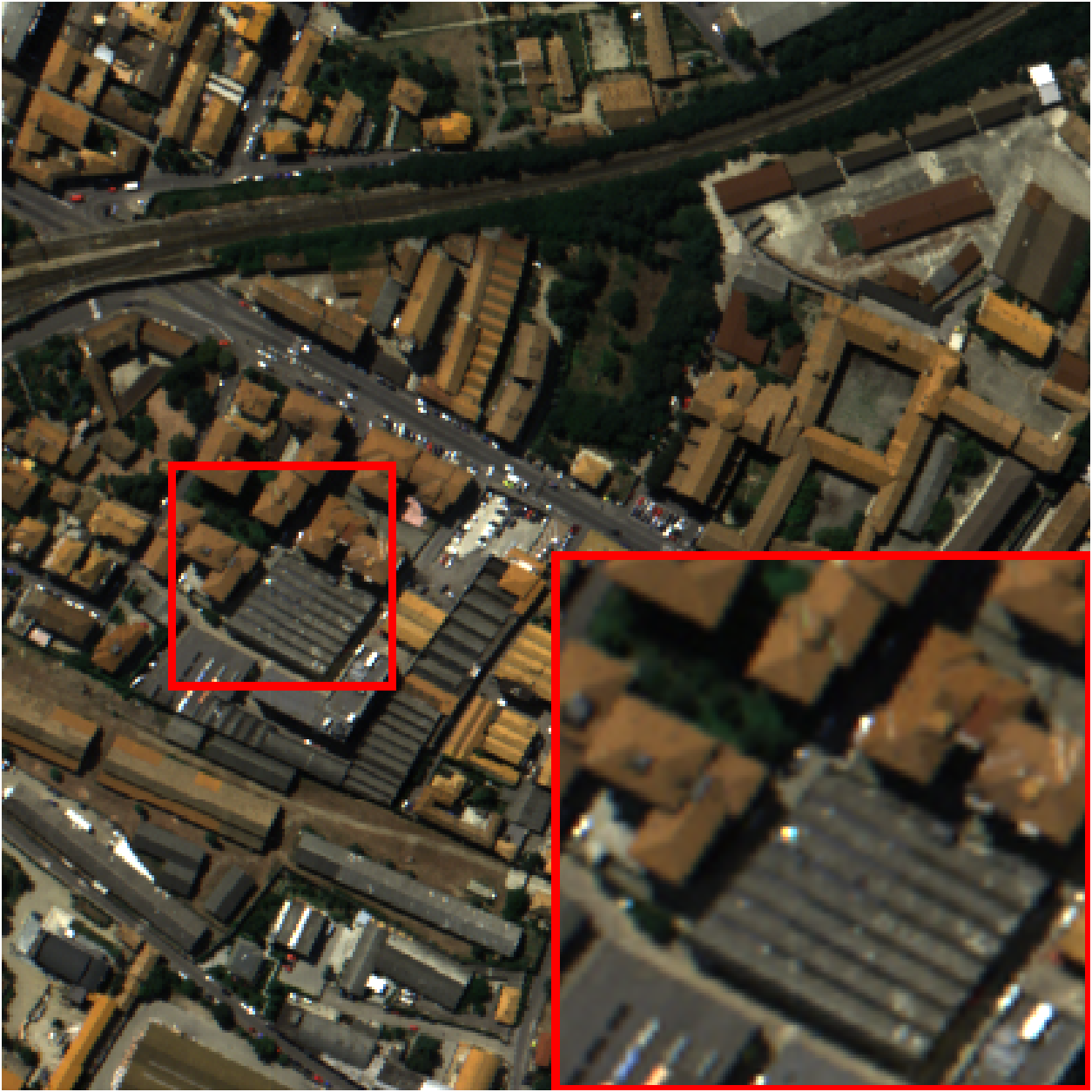} \\
		HRHS
	\end{minipage}
	\begin{minipage}[t]{\mysize}
		\centering
		\fontsize{10}{11}\selectfont
		\includegraphics[width=\mysize]{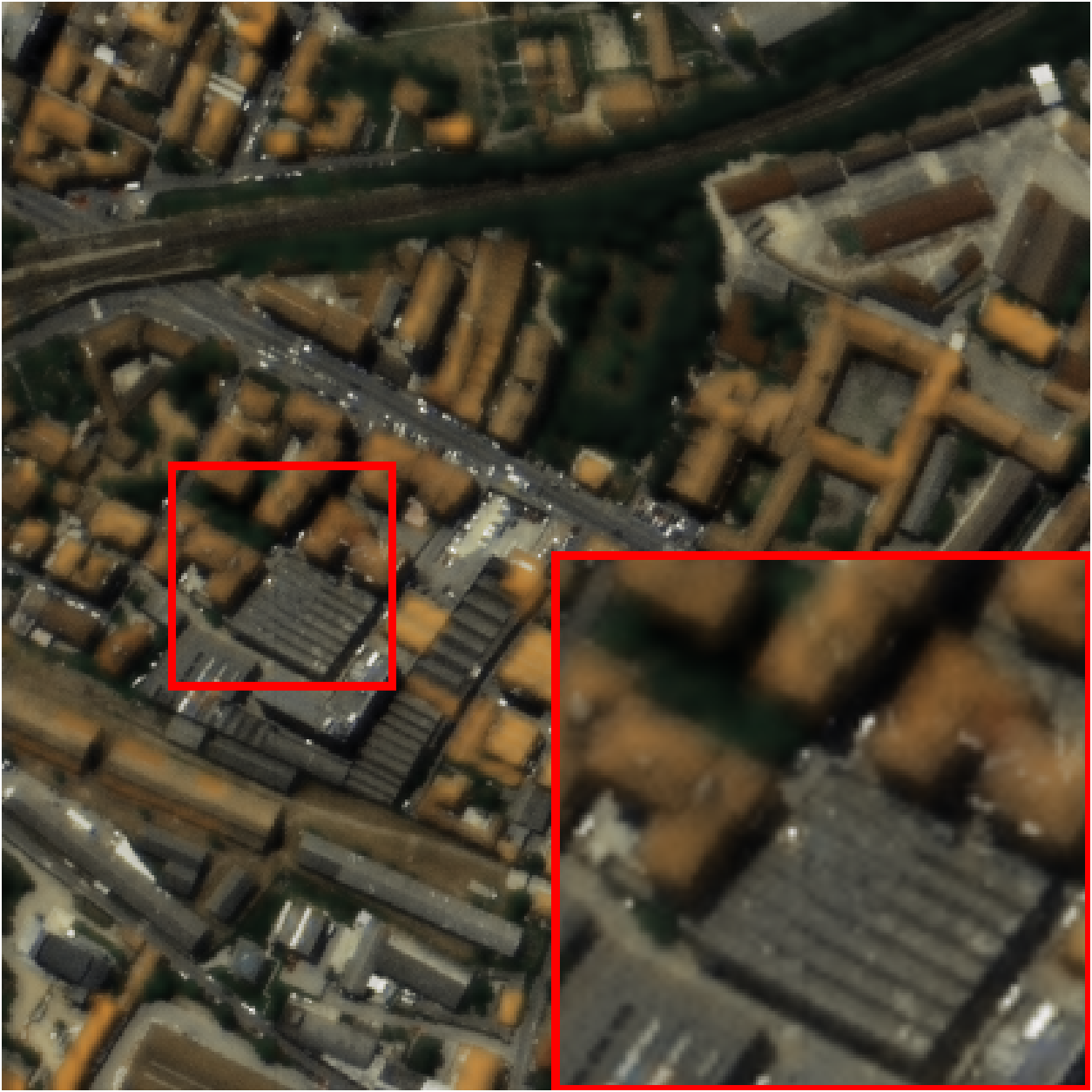} \\
		CNMF
	\end{minipage}
	\begin{minipage}[t]{\mysize}
		\centering
		\fontsize{10}{11}\selectfont
		\includegraphics[width=\mysize]{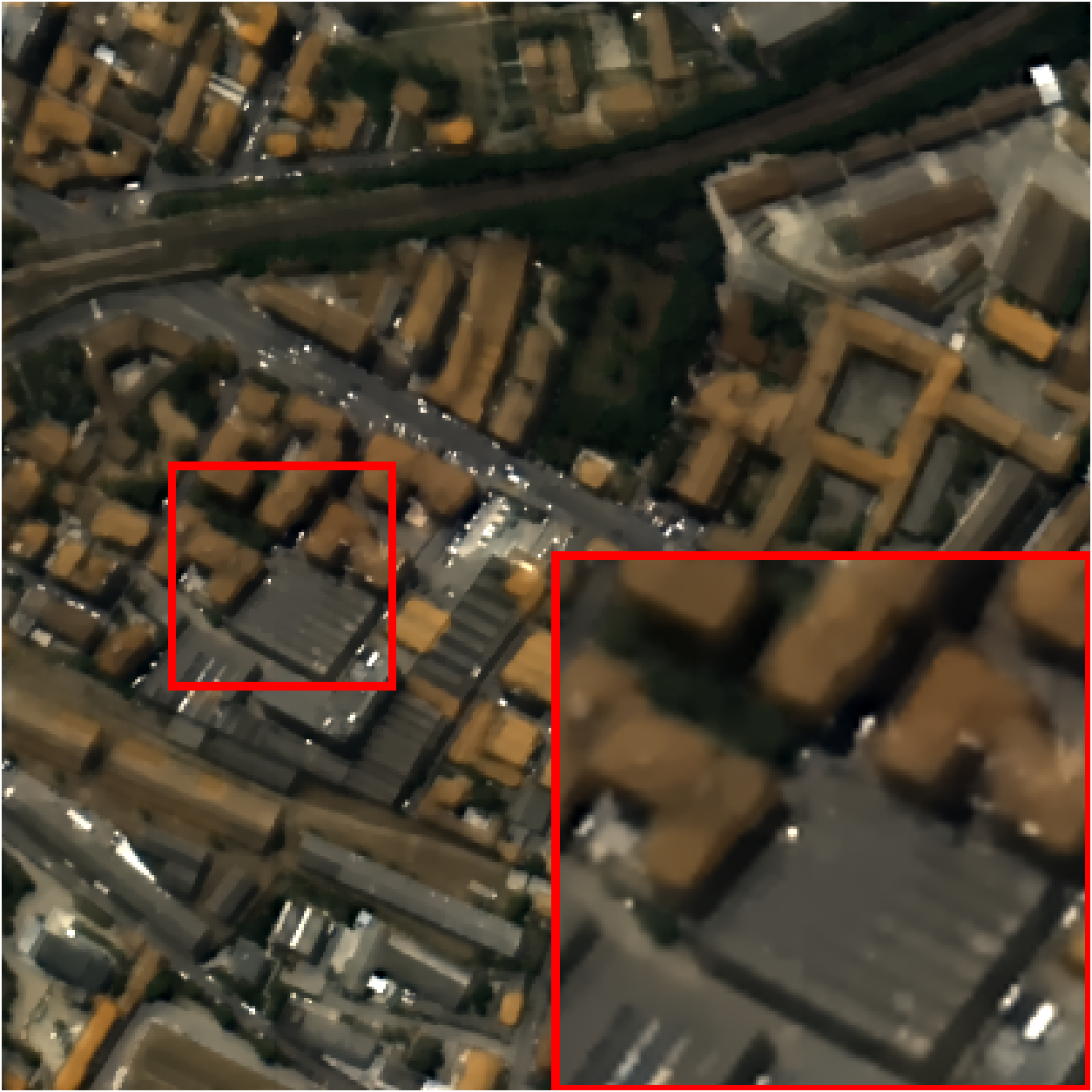} \\
		HySure
	\end{minipage}
	\begin{minipage}[t]{\mysize}
		\centering
		\fontsize{10}{11}\selectfont
		\includegraphics[width=\mysize]{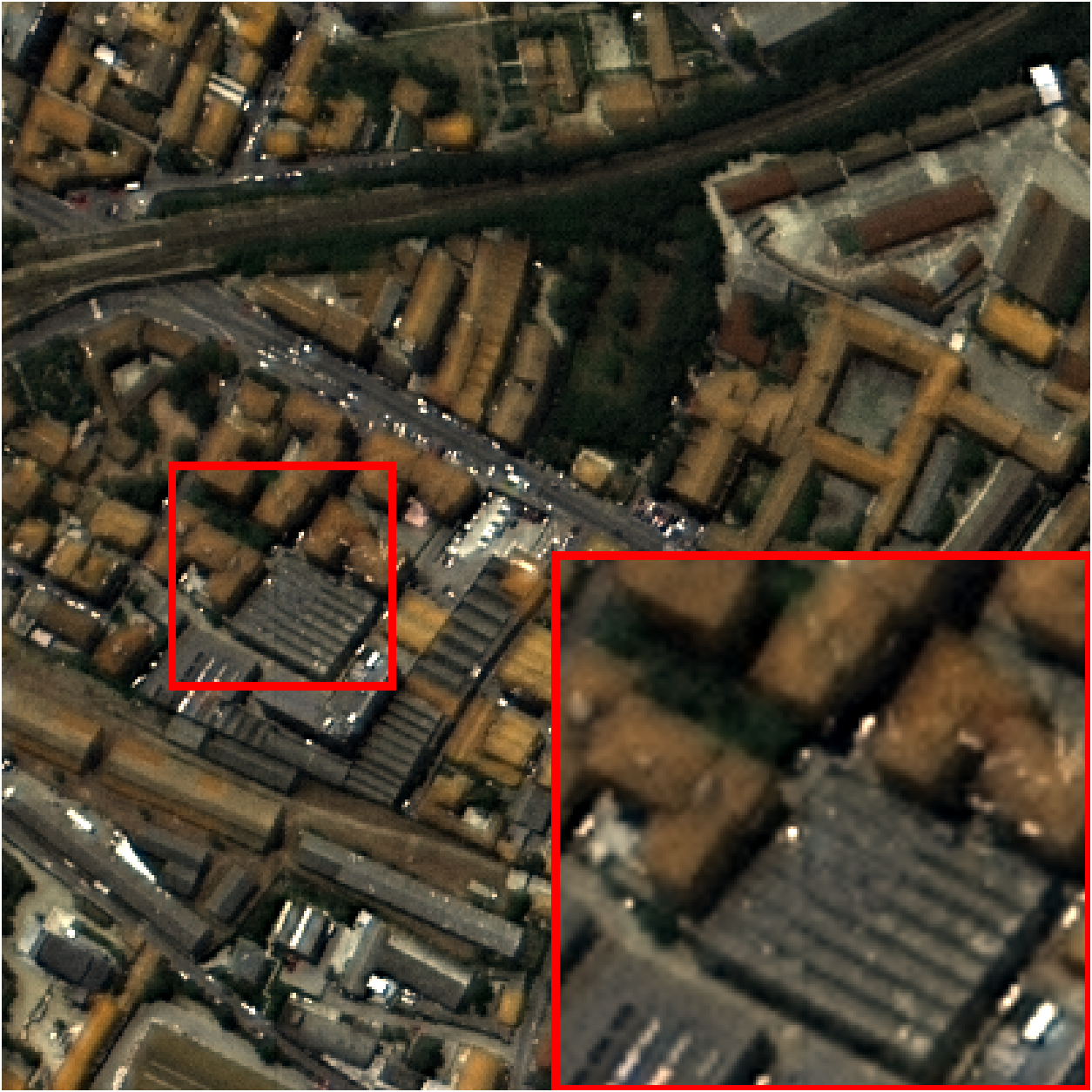} \\
		GLP
	\end{minipage} \vspace{3pt} \\

	\begin{minipage}[t]{\mysize}
		\centering
		\fontsize{10}{11}\selectfont
		\includegraphics[width=\mysize]{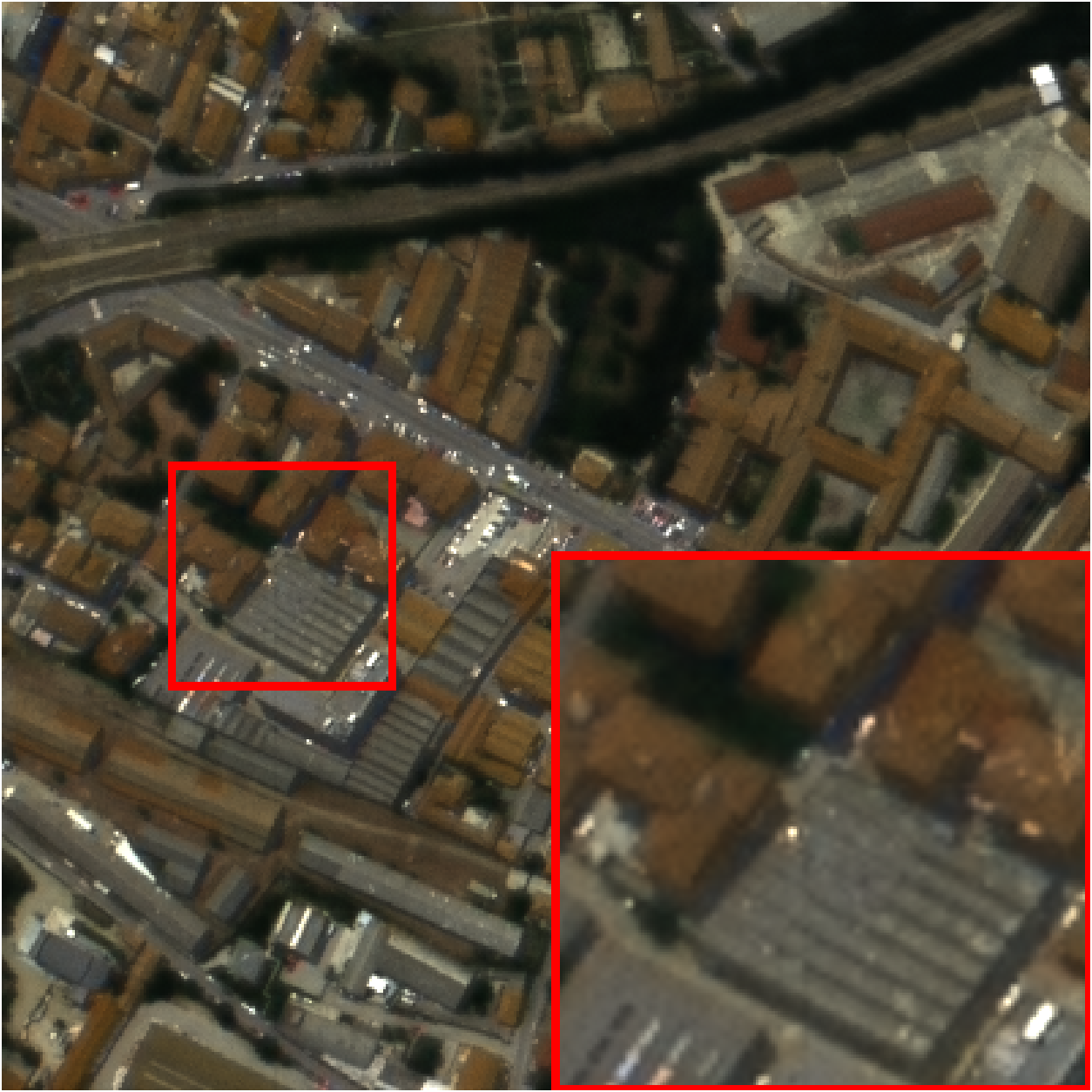} \\
		Brovey
	\end{minipage}
	\begin{minipage}[t]{\mysize}
		\centering
		\fontsize{10}{11}\selectfont
		\includegraphics[width=\mysize]{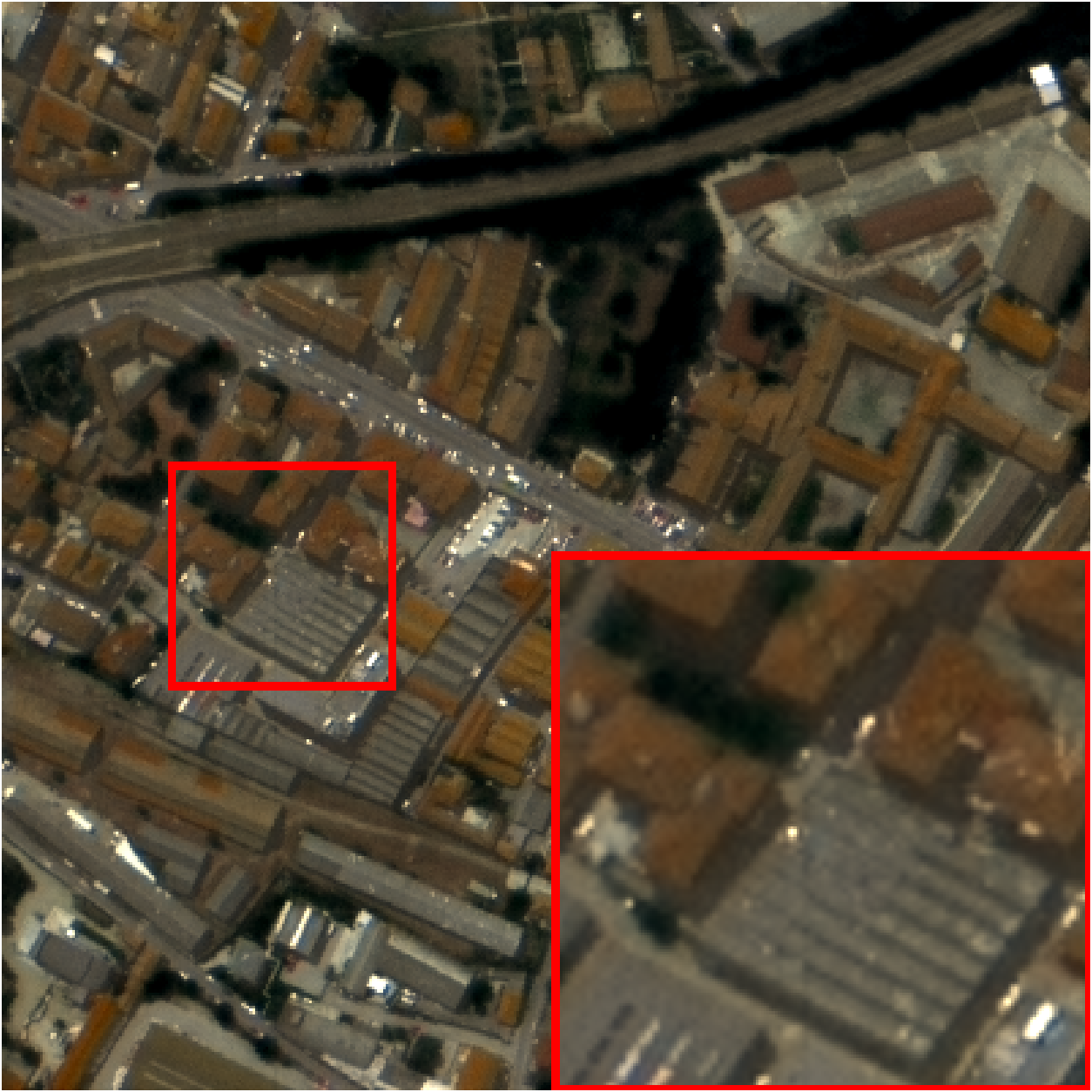} \\
		GS
	\end{minipage}	
	\begin{minipage}[t]{\mysize}
		\centering
		\fontsize{10}{11}\selectfont
		\includegraphics[width=\mysize]{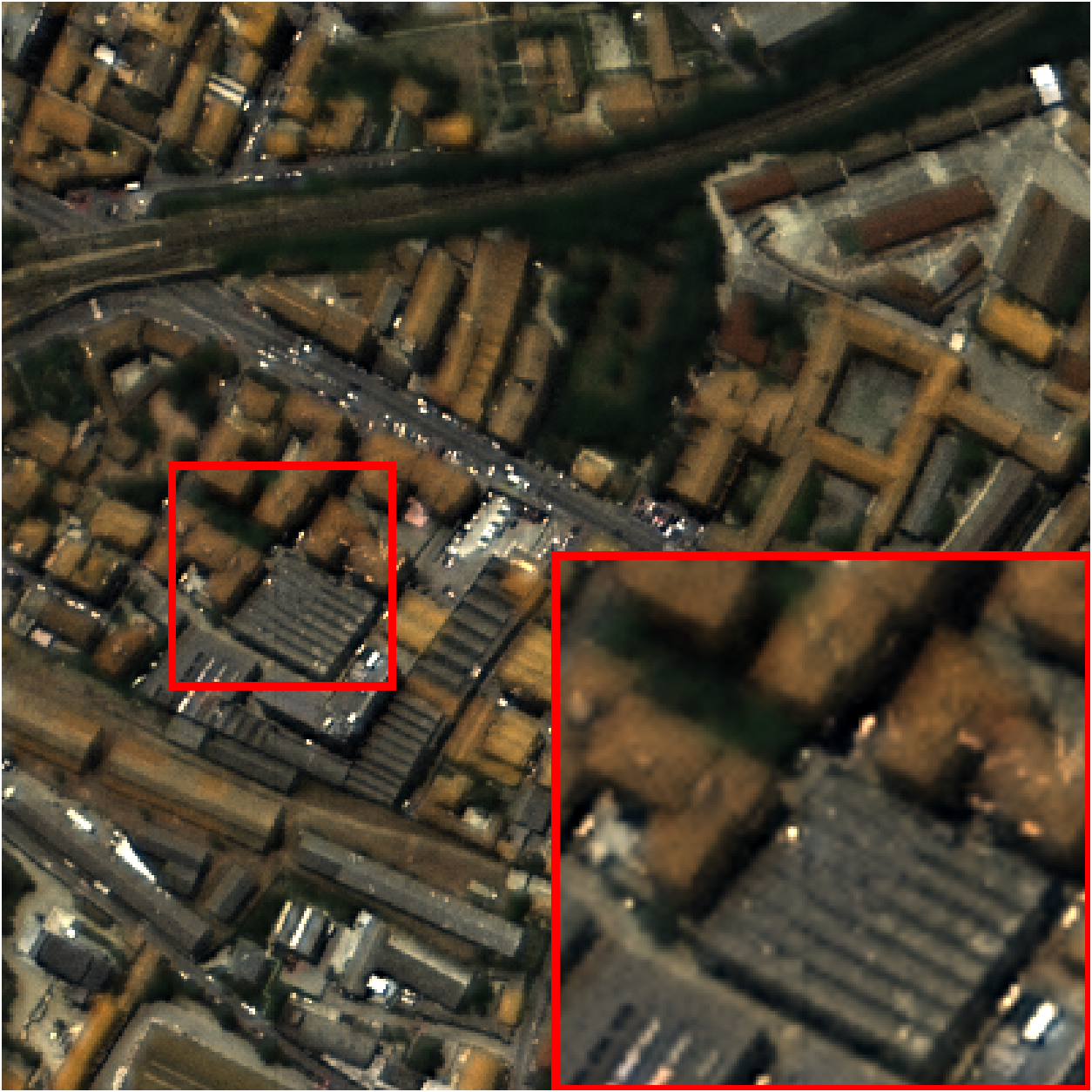} \\
		AWLP
	\end{minipage}
	\begin{minipage}[t]{\mysize}
		\centering
		\fontsize{10}{11}\selectfont
		\includegraphics[width=\mysize]{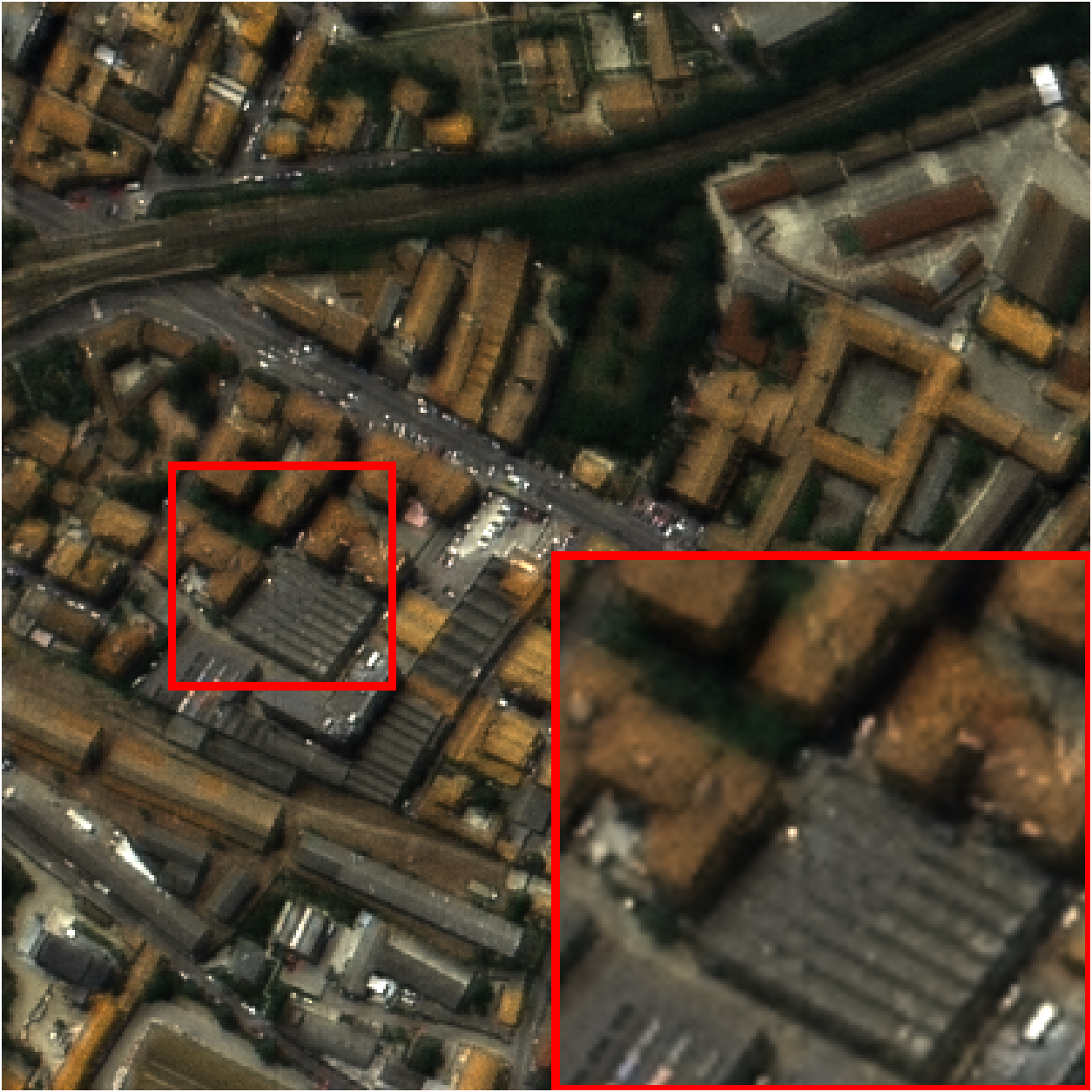} \\
		SFIM 
	\end{minipage} \vspace{3pt}\\
	\begin{minipage}[t]{\mysize}
		\centering
		\fontsize{10}{11}\selectfont
		\includegraphics[width=\mysize]{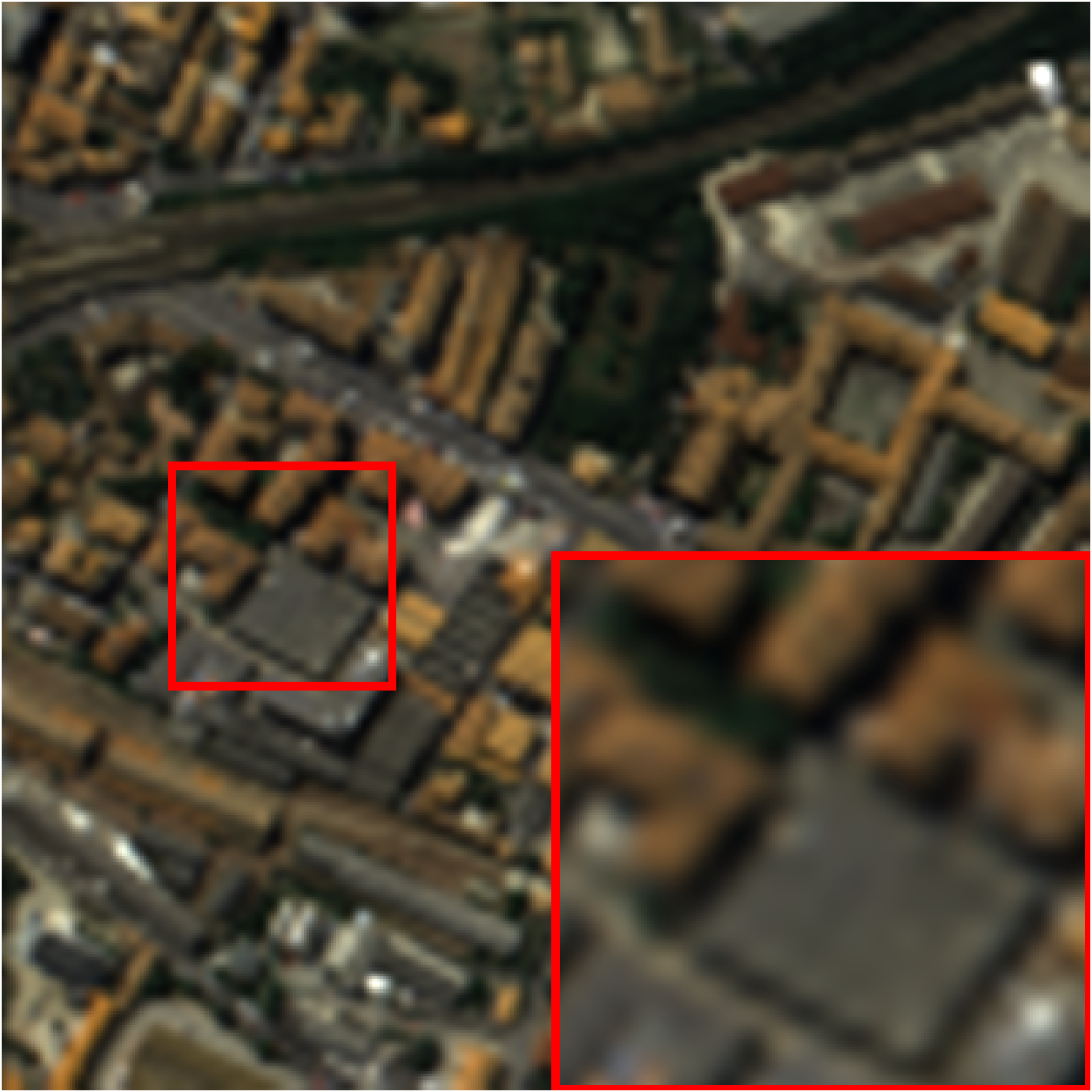} \\
		FUSE
	\end{minipage}
	\begin{minipage}[t]{\mysize}
		\centering
		\fontsize{10}{11}\selectfont
		\includegraphics[width=\mysize]{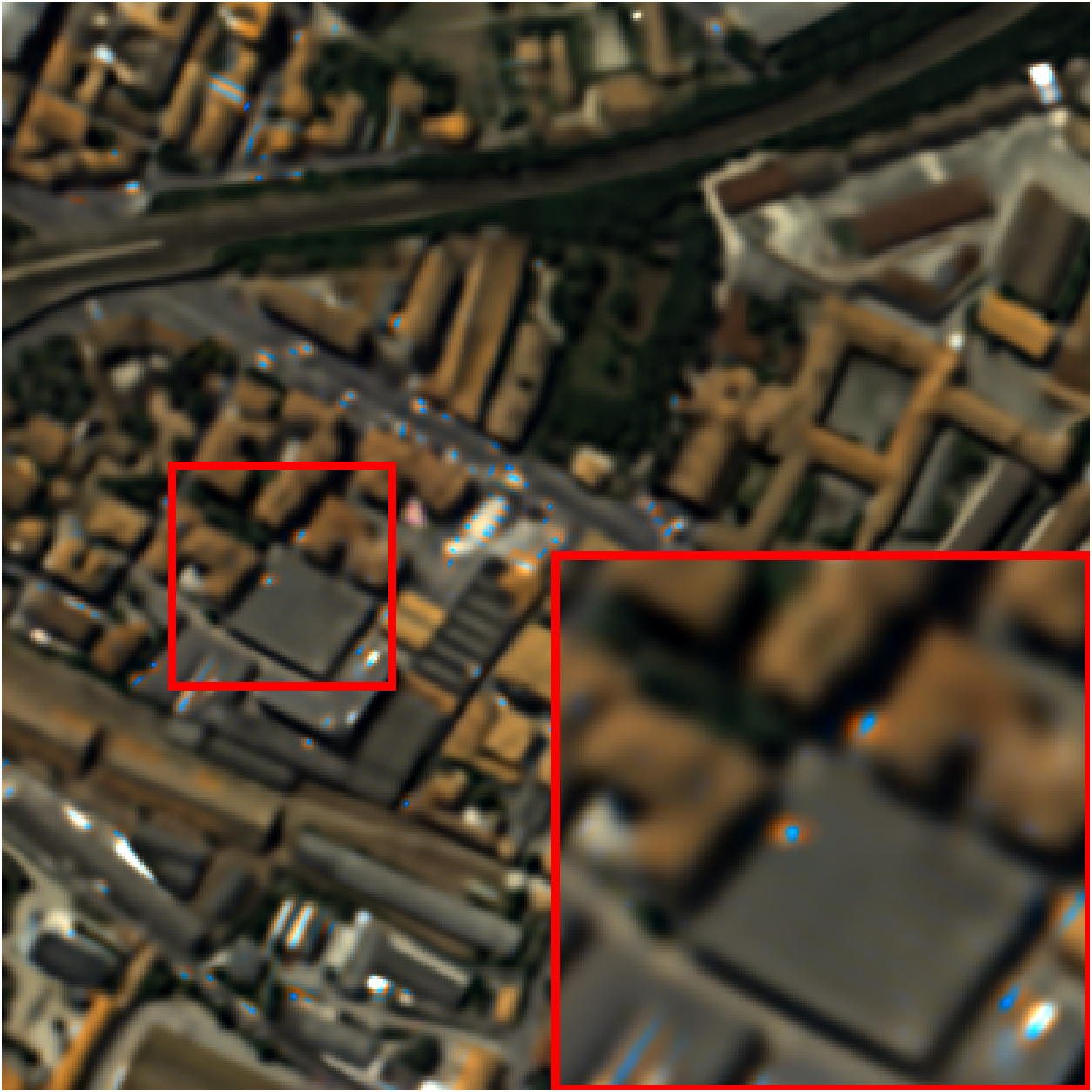} \\
		CNNFUS
	\end{minipage}
	\begin{minipage}[t]{\mysize}
		\centering
		\fontsize{10}{11}\selectfont
		\includegraphics[width=\mysize]{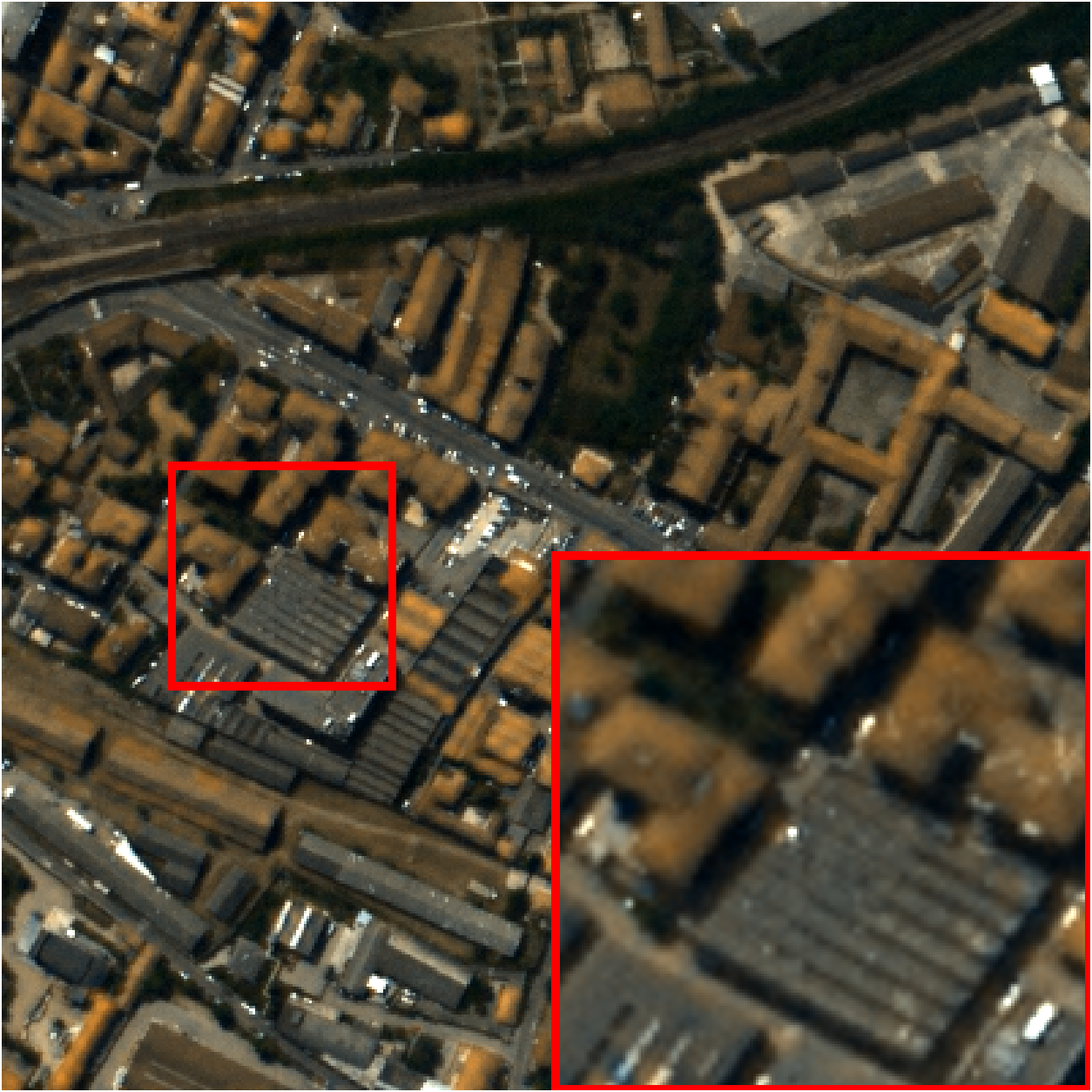} \\
		Ours
	\end{minipage}
	\caption{Visual results on Pavia dataset. Shown bands are [20,40,60].}
	\label{fig-pav}
\end{figure}

\begin{figure}[ht]
	\newcommand{\mysize}{5cm}
	\centering
	\begin{minipage}[t]{\mysize}
		\centering
		\fontsize{10}{11}\selectfont
		\includegraphics[width=\mysize]{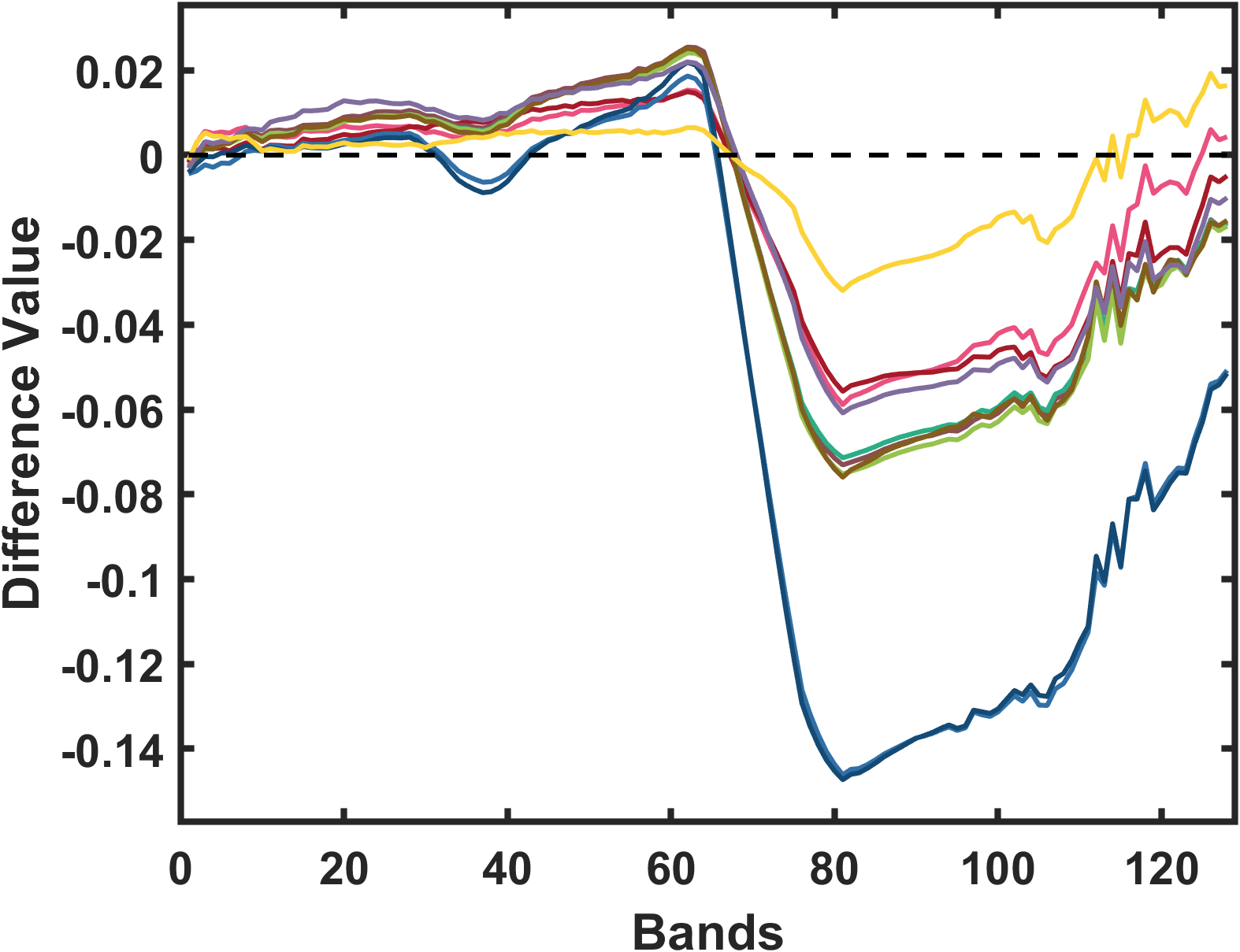} \\
		~~~~~Chikusei (30, 40)
	\end{minipage} \hspace{4pt}
	\begin{minipage}[t]{\mysize}
		\centering
		\fontsize{10}{11}\selectfont
		\includegraphics[width=\mysize]{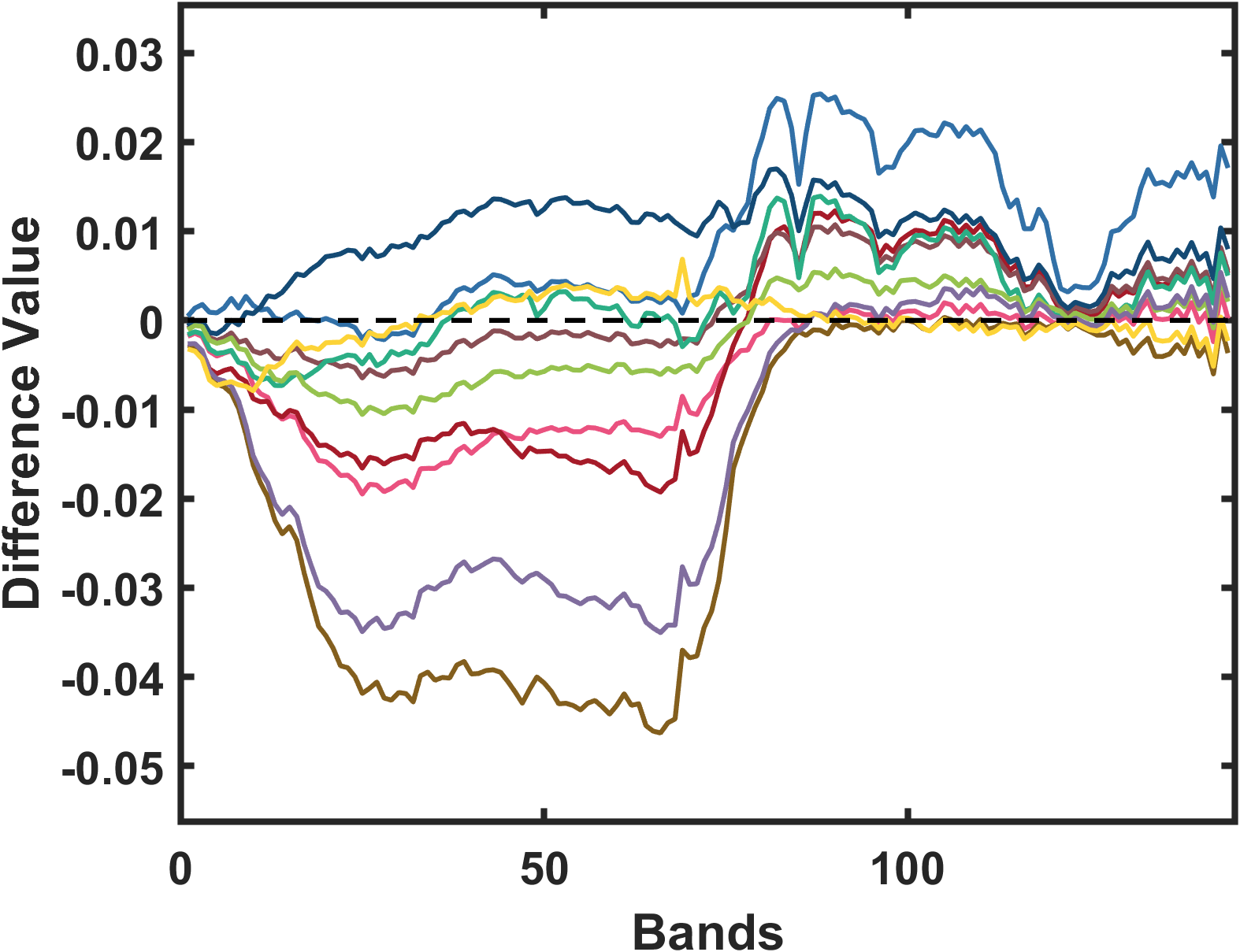} \\
	    ~~~~~~Houston (50, 70)
	\end{minipage} \vspace{4pt} \\
	\begin{minipage}[t]{\mysize}
		\centering
		\fontsize{10}{11}\selectfont
		\includegraphics[width=\mysize]{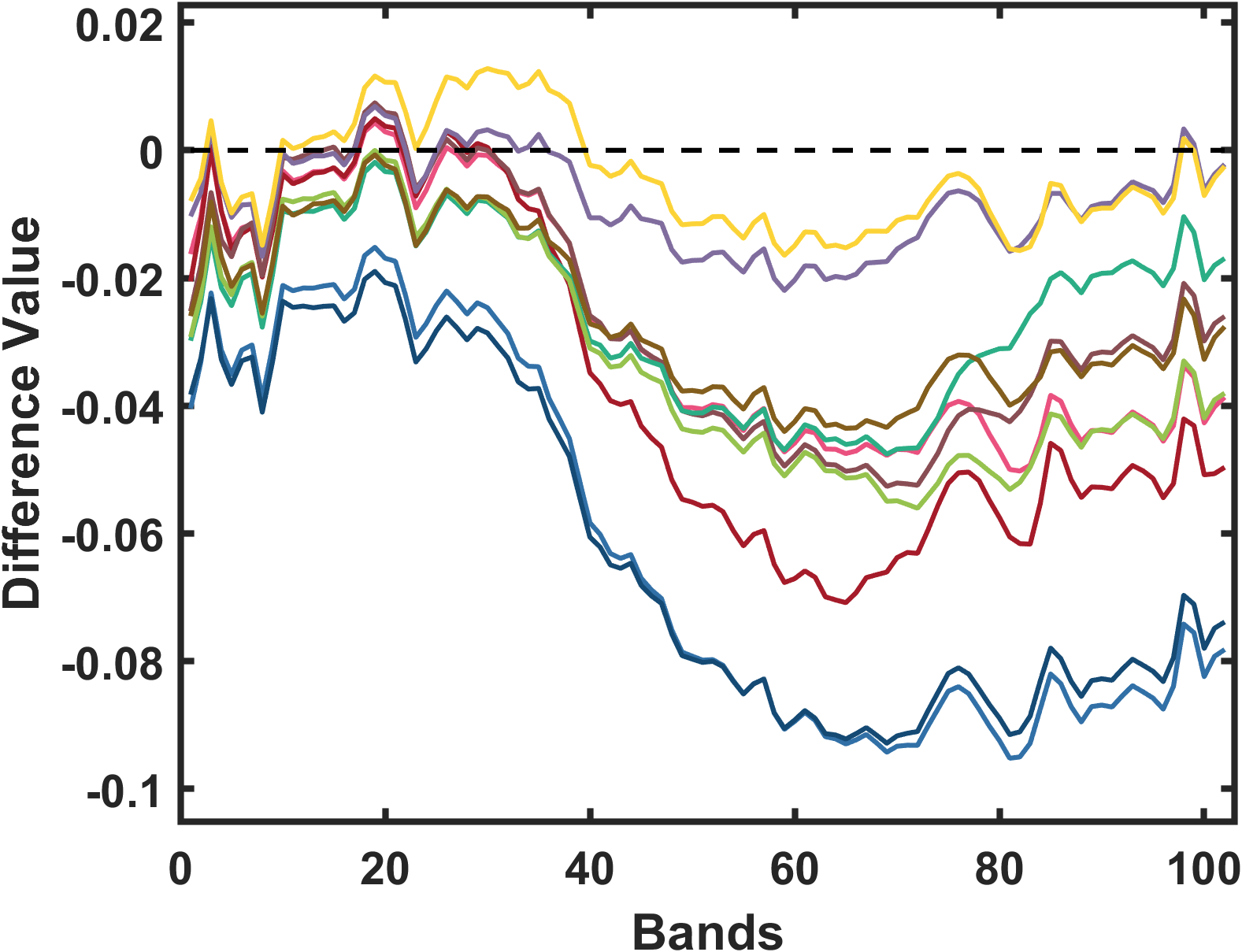} \\
		~~~~~~Pavia (160, 300)
	\end{minipage} \hspace{4pt}
	\begin{minipage}[t]{\mysize}
		\centering
		\fontsize{10}{11}\selectfont
		\includegraphics[width=2cm]{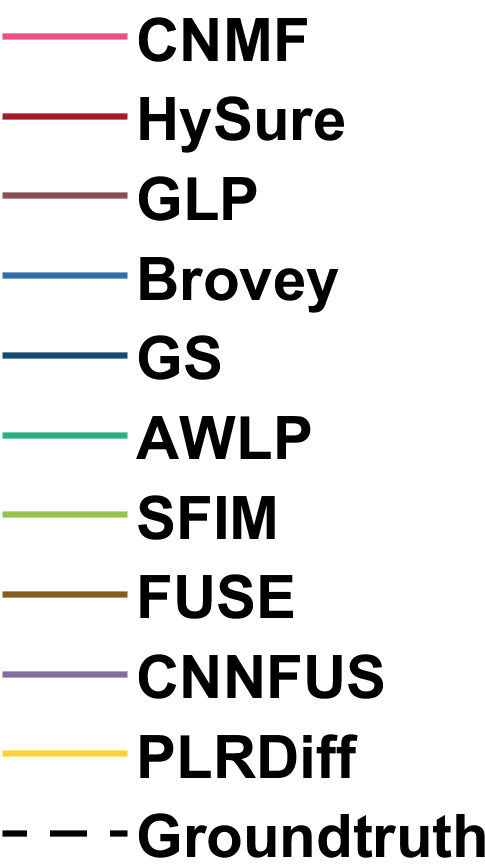} \\
	\end{minipage}
	\caption{Difference value between the HRHS image and the images restored by all competitive methods. Pixel positions for the Chikusei, Houston and Pavia datasets are (30, 40), (50, 70) and (160, 300), respectively.}
\label{fig-sscurve}
\end{figure}

\subsection{Experiment Results}
In this section, we compare the proposed method with 9 popular methods, including CNMF \cite{CNMF}, HySure \cite{HySure}, GLP \cite {MTF-GLP}, Brovey \cite{Brovey}, GS \cite{GS}, AWLP \cite{AWLP}, SFIM \cite{SFIM}, FUSE \cite{Bayesian-fusion-of-multi-band-images} and CNNFUS \cite{CNNFUS}. Specifically, Brovey and GS are CS-based methods. GLP, SFIM and AWLP are MRA-based methods. CNMF, HySure and FUSE are Bayesian methods. CNNFUS can also be considered as a Bayesian method where a deep denoiser is used to capture the image prior.

Table \ref{tab-chikusei} - Table \ref{tab-pavia} reports the quantitative results of the comparison methods and the proposed PLRDiff on the testing datasets. We can see that, overall, the proposed method gains superior performance to other competitive methods. From the values of PSNR, Q2N and SCC, it can be seen that the proposed PLRDiff can well restore the spatial details. This means that the utilized pre-train diffusion model has effectively aligned the low-rank subspace and generated the base tensor. Meanwhile, the spectral information retrieved by our method is also remarkable, showing the effectiveness of the proposed low-rank representation technique and the way to estimate the coefficient matrix. Additionally, the proposed method can flexibly handle HS images with different spectral dimensions due to the usage of low-rank factorization. 

The visual comparison results (pseudo-color images) of all the competing methods are shown in Fig. \ref{fig-chi} - Fig. \ref{fig-pav}. For the Chikusei dataset, we can easily observe from Fig. \ref{fig-chi} that some competitive methods contain apparent blurring effects and can not effectively restore the details. For the Houston dataset, the proposed method obtains sharper details than other competing methods as shown in Fig. \ref{fig-hou}. For the Pavia dataset shown in Fig. \ref{fig-pav}, the color distortion of the restored images by some competitive methods shows that these images loss the spectral information. The proposed method performs well both in characterizing spatial details and preserving spectral information. In Fig. \ref{fig-sscurve}, we show the difference value between the HRHS image and the restored images along the spectral dimension. It can be seen that the proposed PLRDiff could finely reconstruct the spectral information.

\begin{figure}[t]
	\newcommand{\mysize}{2.55cm}
	\newcommand{\wid}{3.5cm}
	\centering
	\subfloat[Chikusei]{
	\begin{minipage}[t]{\wid}
		\centering
		\fontsize{10}{11}\selectfont
		\includegraphics[width=\wid]{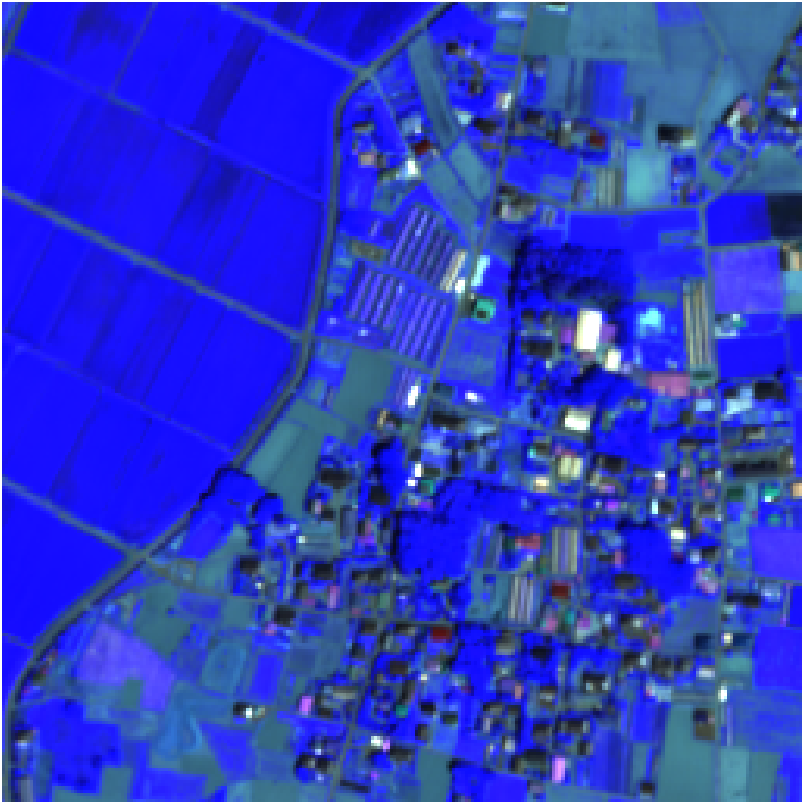}\\
		$\A$ of HRHS \vspace{5pt}\\
		\includegraphics[width=\wid]{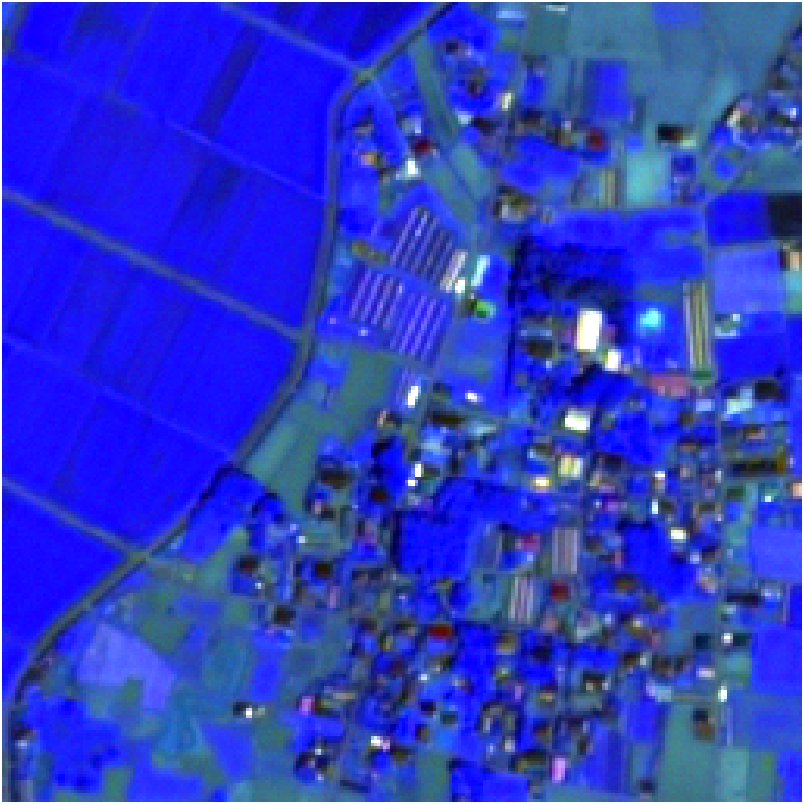}\\
		predicted $\A$ 
	\end{minipage}}\hspace{2pt}
    \subfloat[Houston]{
    \begin{minipage}[t]{\wid}
   		\centering
   		\fontsize{10}{11}\selectfont
   		\includegraphics[width=\wid]{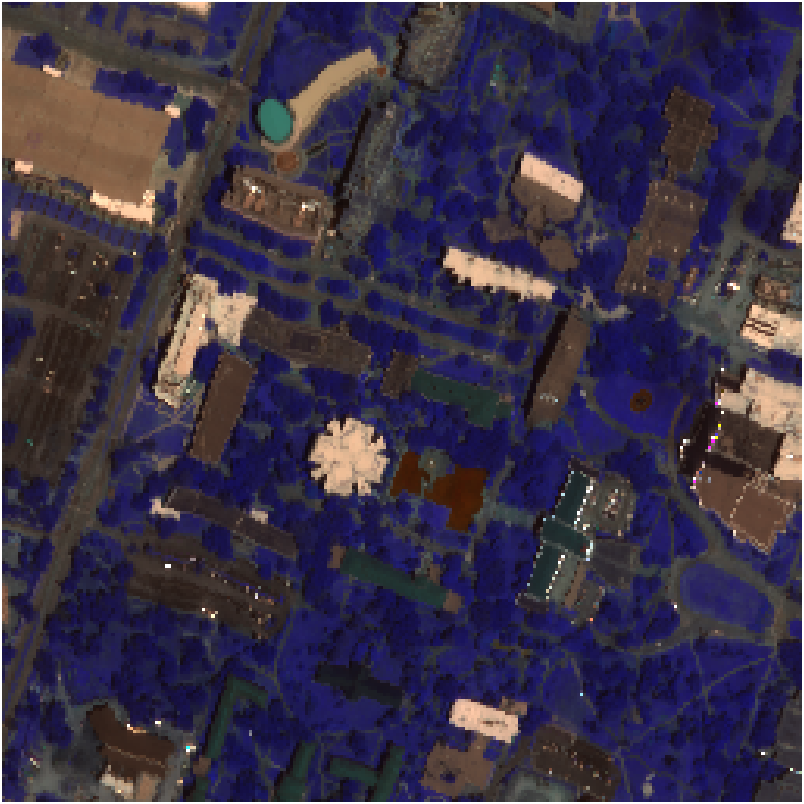}\\
   		$\A$ of HRHS \vspace{4pt}\\
   		\includegraphics[width=\wid]{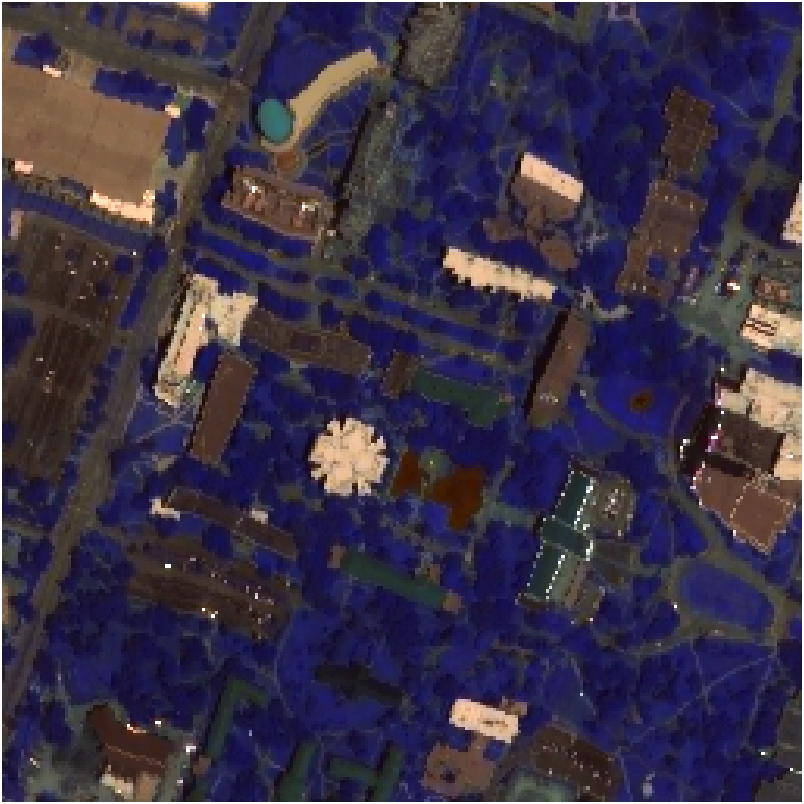}\\
   		predicted $\A$ 
    \end{minipage}}\hspace{2pt}
    \subfloat[Pavia]{
    \begin{minipage}[t]{\wid}
   		\centering
   		\fontsize{10}{11}\selectfont
   		\includegraphics[width=\wid]{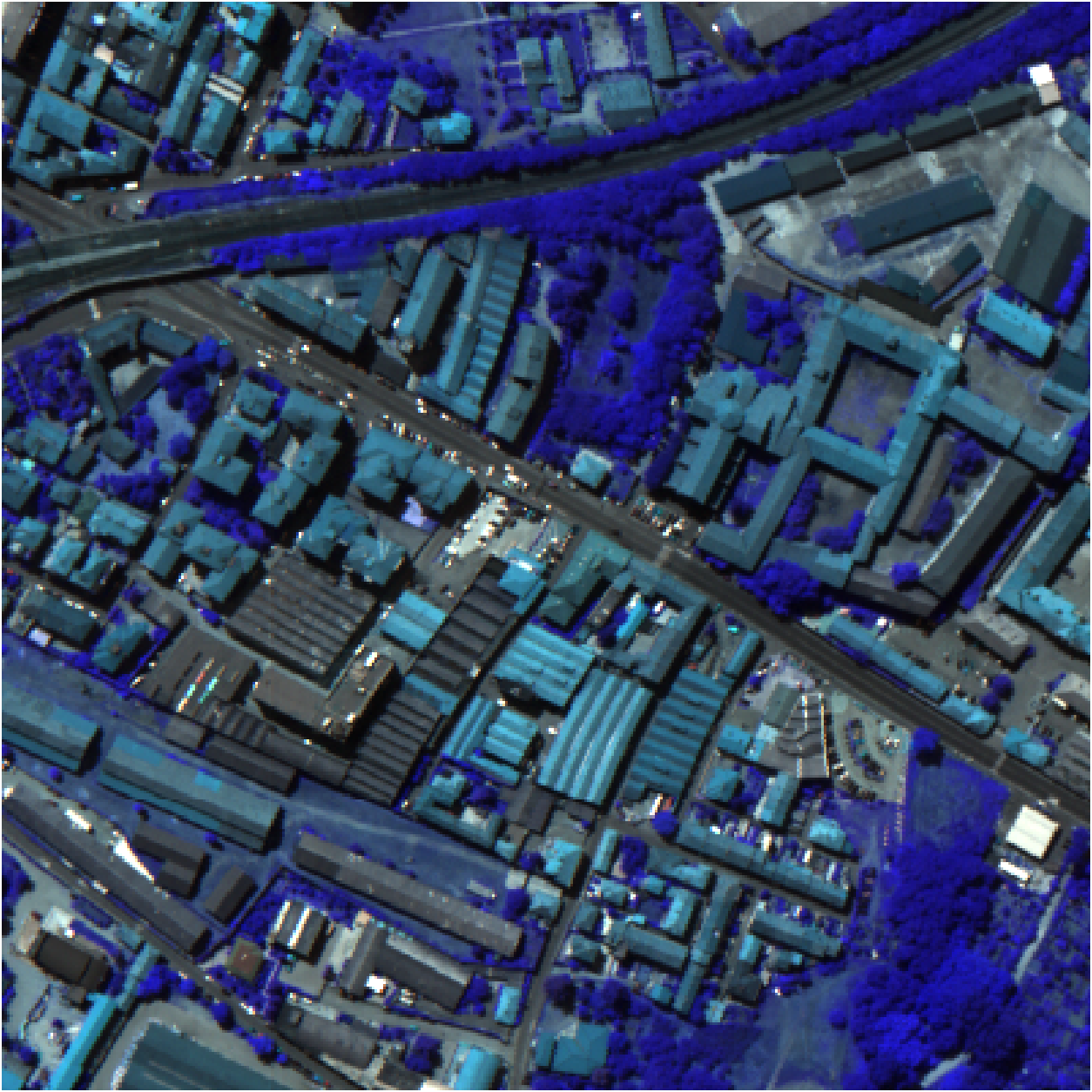}\\
   		$\A$ of HRHS \vspace{4pt}\\
   		\includegraphics[width=\wid]{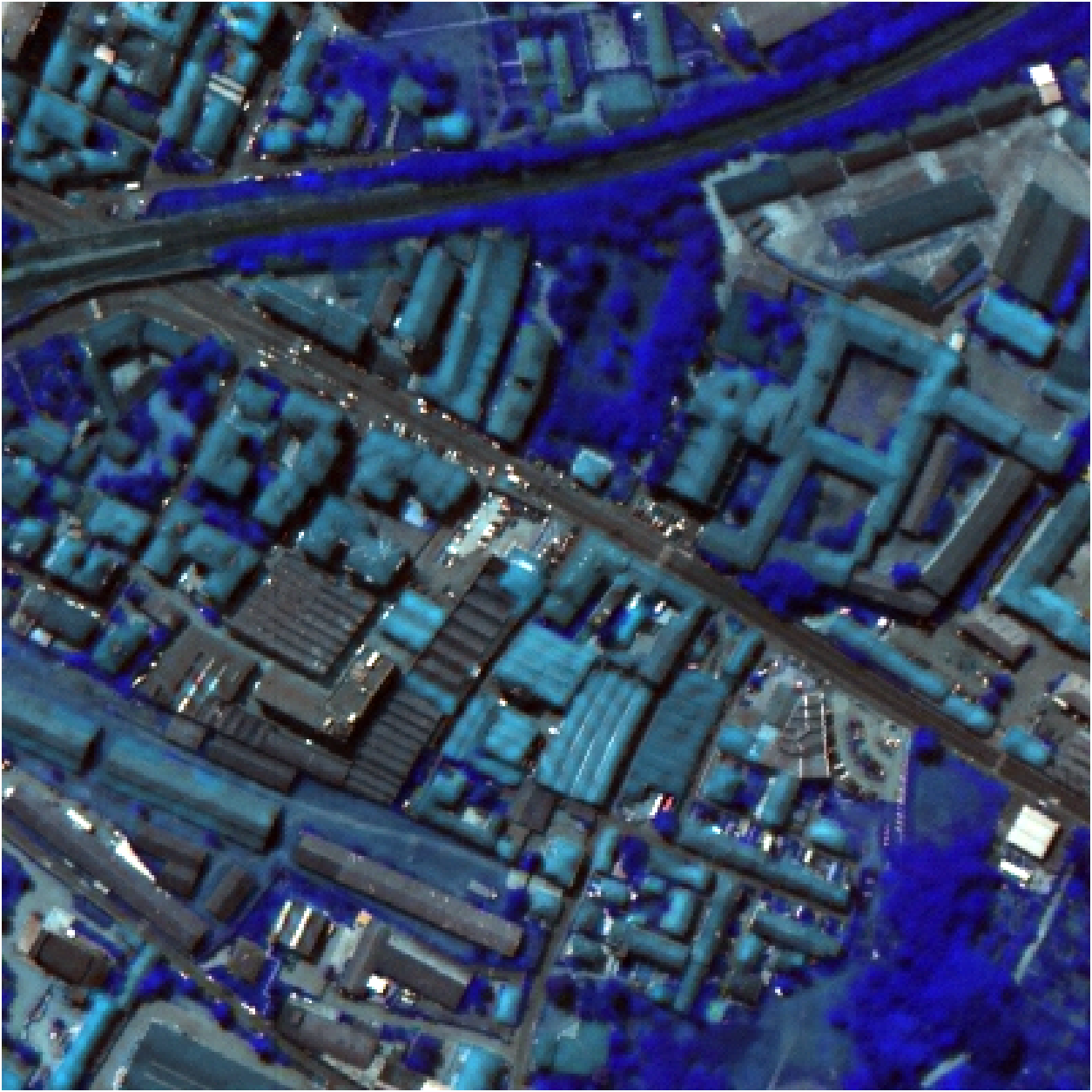}\\
   		predicted $\A$ 
    \end{minipage}}
	\caption{Comparison of base tensor $\A$ predicted by our method and extracted from the HRHS image.}
	\label{fig-A}
\end{figure}

Besides, we also visualize the predicted base tensor $\A$ on the Chikusei, Houston and Pavia datasets by our method in Fig. \ref{fig-A}. As can be seen, the spatial and spectral information are well preserved in the predicted base tensor. Thus, the proposed method could finely utilize a pre-trained diffusion model to reconstruct the base tensor.  

\subsection{Comparing with DL-based methods}
In this section, we compare the proposed PLRDiff with two DL-based methods, i.e., FusionNet \cite{FusionNet} and DSNet \cite{A-deep-shallow-fusion-network-with-multidetail-extractor-and-spectral-attention-for-hyperspectral-pansharpening}. The two networks are trained under supervised manner, where network parameters are optimized by minimizing the distance between the true HRHS image and the network output. The training dataset of Pavia\footnote{https://github.com/liangjiandeng/HyperPanCollection} is used to train both networks. Codes are publicly available on their project page. Then the trained networks are used for the three testing datasets. 

The Root Mean Square Error (RMSE) results are presented in Table \ref{tab-net}. We can see that, for the Pavia dataset, the performance of PLRDiff is comparable with the two networks. For the Chikusei dataset and the Houston dataset, the performance of the network methods drops rapidly because of the distribution shift between the training and testing datasets. However, the proposed PLRDiff still performs well. 

\begin{table}[ht]
	\renewcommand{\arraystretch}{1.25}
	\newcommand{\mysize}{1.6cm}
	\fontsize{10}{11}\selectfont
	\caption{RMSE values of FusionNet, DSNet and PLRDiff on the test datasets. FusionNet and DSNet are trained using the training dataset of Pavia. The best results are in \textbf{bold}.}
	\label{tab-net}
	\centering
	\begin{tabular}{ M{2cm} | M{\mysize} M{\mysize} M{\mysize}}
		\Xhline{0.8pt}
		methods & Pavia & Chikusei & Houston \\
		\hline 
		FusionNet & 0.0345 & 0.0961 & 0.0587 \\
		DSNet & 0.0332 & 0.1006 & 0.0558 \\
		PLRDiff & \tb{0.0307} & \tb{0.0327} & \tb{0.0140} \\
		\Xhline{0.8pt}
	\end{tabular}
\end{table}

\section{Discussion}\label{sec-discussion}
\subsection{Effects of the Coefficient Matrix}
\begin{figure}[ht]
	\newcommand{\mysize}{4cm}
	\centering
	\begin{minipage}[t]{\mysize}
		\centering
		\fontsize{10}{11}\selectfont
		\includegraphics[width=\mysize]{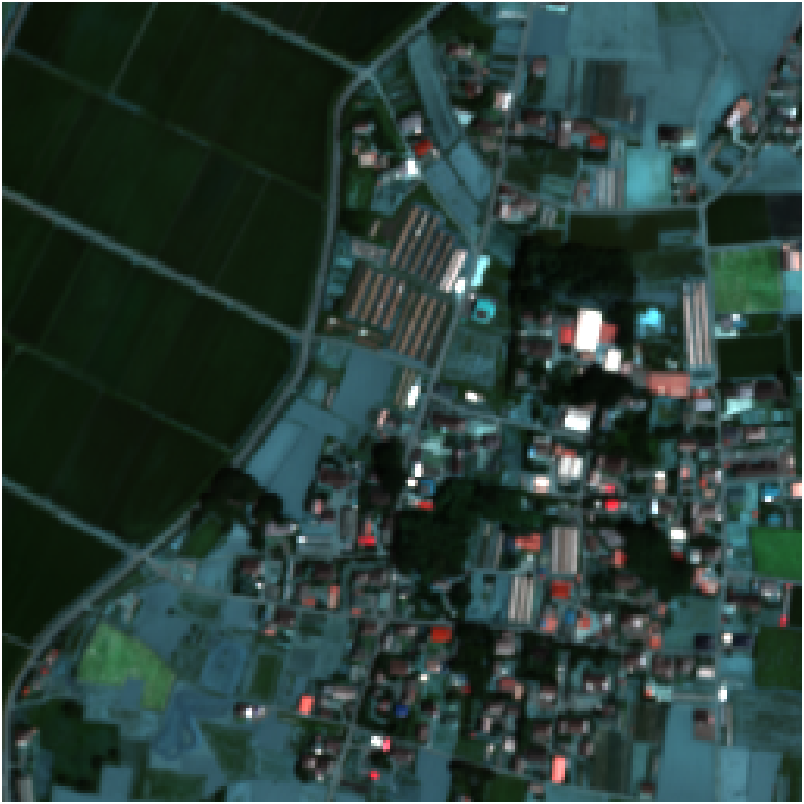} \\
		Chikusei 
	\end{minipage} \hspace{2pt}
	\begin{minipage}[t]{\mysize}
		\centering
		\fontsize{10}{11}\selectfont
		\includegraphics[width=\mysize]{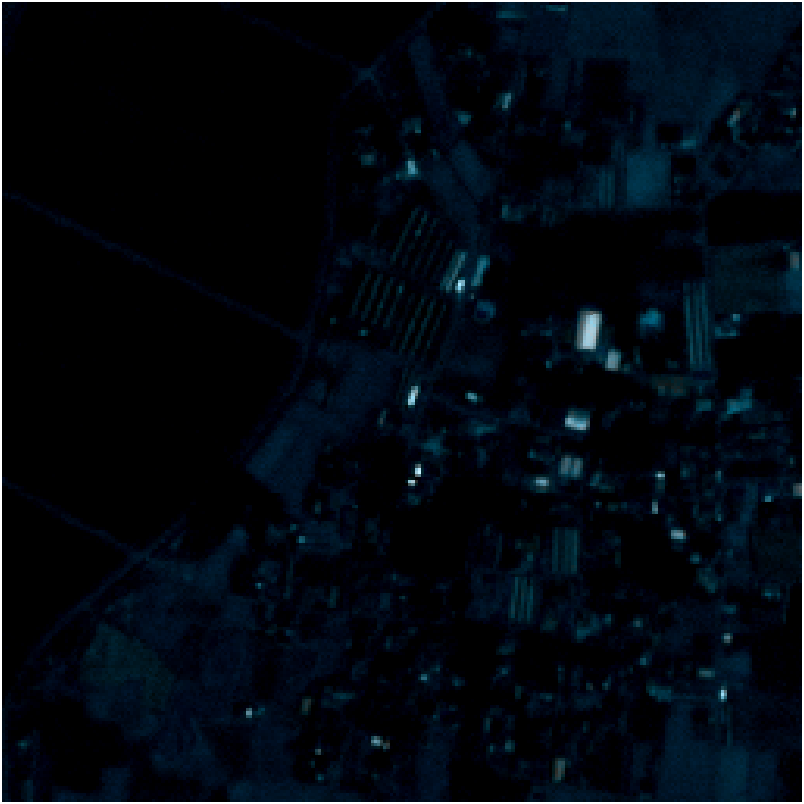}\\
		PLRDiff-SVD
	\end{minipage}
	\begin{minipage}[t]{\mysize}
		\centering
		\fontsize{10}{11}\selectfont
		\includegraphics[width=\mysize]{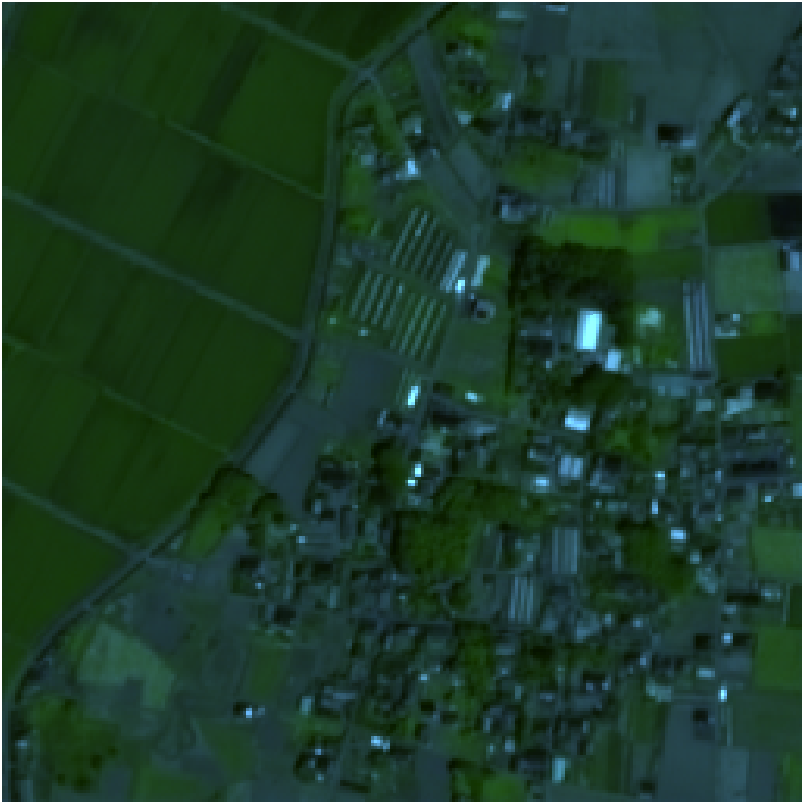}\\
		PLRDiff-Rand
	\end{minipage} 
	\caption{Pseudo-color images of the Chikusei dataset and the corresponding restored HRHS images by PLRDiff-SVD and PLRDiff-Rand.}
	\label{fig-svdrand}
\end{figure}

In Sec. \ref{sec-low-rank-factorization}, since the pre-trained diffusion model is assumed to learn the image distribution, we construct the base tensor $\A$ by the slices of the HRHS image itself. Then we estimate the coefficient matrix $E$ by (\ref{estimate-E}). Actually, if the distribution of the base tensor does not necessarily match that learnt by the diffusion model, the spectral information could be severely damaged for the restored HRHS image. To illustrate this, we choose two other ways to set the coefficient matrix $E$. One of them is through SVD. Suppose $U\Sigma V^T$ is the SVD form of $\brm{Y}_{(3)}$, then we set $E=U[:, 1:s]$ which means the first $s$ columns of $U$. The other one is randomly sampling $E$ from the uniform distribution supported on $[0,1]$. Then we apply the PLRDiff algorithm based on the two settings of $E$. The algorithm with $E$ that is calculated by SVD is denoted by PLRDiff-SVD and the algorithm with $E$ that is randomly selected is denoted by PLRDiff-Rand. In Fig. \ref{fig-svdrand}, we show the pseudo-color images of the results by the two settings. It shows that the diffusion model is able to help to reconstruct some of the spatial information although the features of the base tensor deviate from the diffusion model. However, the spectral information is destroyed. This is because the non-matching features of the true base tensor and the diffusion model are conflicting with the coefficient matrix $E$. Beyond the factorization form used in this work, it is also possible to project the HS image into other a lower-dimensional subspace and train a diffusion model to learn the distribution of the projected images. We will investigate this issue in our future research.

\subsection{Total diffusion steps}
In this section, we verify the performance and running time of the proposed PLRDiff  under different diffusion steps $T$ using Chikusei validation dataset. The results are presented in Table \ref{tab-steps}. For the proposed method, we can see that increasing the sampling steps could bonus the performance because the discretization of the SDE is more compact. Simultaneously, the runing time also increases as the sampling steps increases. And the marginal benefits is gradually diminished. To balance the performance and the running time, we select the diffusion steps as 500.

\begin{table}[ht]
	\renewcommand{\arraystretch}{1.25}
	\newcommand{\mysize}{1.1cm}
	\fontsize{10}{11}\selectfont
	\caption{Performance and running time of PLRDiff under different total numbers of diffusion steps $T$.}
	\label{tab-steps}
	\centering
	\begin{tabular}{ M{2cm} | M{\mysize} M{\mysize} M{\mysize} M{\mysize} M{\mysize} M{\mysize} M{\mysize}}
		\Xhline{0.8pt}
		steps $T$ & 50 & 100 & 300 & 500 & 800 & 1000 & 1500 \\
		\hline 
		Times (s) & 12 & 25 & 73 & 123 & 197 & 246 & 369\\
		PSNR & 23.06 & 28.82 & 32.04 & 32.13 & 32.20 & 32.23 & 32.30 \\
		\Xhline{0.8pt}
	\end{tabular}
\end{table}

\section{Conclusion}\label{conclusion}
In this work, we propose an unsupervised low-rank diffusion method for HS pansharpening by utilizing the low-rank subspace representation technique and pre-trained unconditional diffusion model. We project the HRHS image into a subspace based on the HS image spectral low-rank property. The subspace is designed to be contained in the image field so that it is consistent with the distribution learnt by the pre-trained diffusion model. We design an efficient strategy to calculate the coefficient matrix corresponding to the projection from the LRHS image. Then the pre-trained diffusion model to utilized to sample from the subspace conditioned on the LRHS image, PAN image and the coefficient matrix. Extensive experiments on different datasets show the flexibility, excellent performance and good generalization ability of our method compared with other traditional methods and DL-based methods.




\bibliographystyle{elsarticle-num} 

\bibliography{reference}

% TeX file.

%\begin{thebibliography}{00}
%%% Text of bibliographic item
%
%\bibitem{}
%
%\end{thebibliography}

\end{document}